\documentclass[lettersize,journal]{IEEEtran}
\usepackage{amsmath,amsfonts}
\usepackage{algorithmic}
\usepackage{algorithm}
\usepackage{array}
\usepackage[caption=false,font=normalsize,labelfont=sf,textfont=sf]{subfig}
\usepackage{textcomp}
\usepackage{stfloats}
\usepackage{url}
\usepackage{verbatim}
\usepackage{graphicx}
\usepackage{cite}
\usepackage{multirow}
\usepackage{booktabs}
\usepackage[pagebackref=true,breaklinks=true,letterpaper=true,colorlinks,bookmarks=false]{hyperref}
\ifCLASSOPTIONcompsoc
  \usepackage[nocompress]{cite}
\else
  \usepackage{cite}
\fi
\ifCLASSINFOpdf
\else
\fi

\usepackage{color}
\newcommand{\wqz}[1]{\textcolor{black}{#1}}
\newcommand{\wqzz}[1]{\textcolor{black}{#1}}

\newcommand{\cyh}[1]{\textcolor{black}{#1}}

\newcommand{\ie}{\textit{i.e.}}
\newcommand{\eg}{\textit{e.g.}}
\newcommand{\Eg}{\textit{E.g.}}

\begin{document}
%
% paper title
% Titles are generally capitalized except for words such as a, an, and, as,
% at, but, by, for, in, nor, of, on, or, the, to and up, which are usually
% not capitalized unless they are the first or last word of the title.
% Linebreaks \\ can be used within to get better formatting as desired.
% Do not put math or special symbols in the title.
\title{Learning Discriminative Features for Crowd Counting}
%
%
% author names and IEEE memberships
% note positions of commas and nonbreaking spaces ( ~ ) LaTeX will not break
% a structure at a ~ so this keeps an author's name from being broken across
% two lines.
% use \thanks{} to gain access to the first footnote area
% a separate \thanks must be used for each paragraph as LaTeX2e's \thanks
% was not built to handle multiple paragraphs
%
%
%\IEEEcompsocitemizethanks is a special \thanks that produces the bulleted
% lists the Computer Society journals use for "first footnote" author
% affiliations. Use \IEEEcompsocthanksitem which works much like \item
% for each affiliation group. When not in compsoc mode,
% \IEEEcompsocitemizethanks becomes like \thanks and
% \IEEEcompsocthanksitem becomes a line break with idention. This
% facilitates dual compilation, although admittedly the differences in the
% desired content of \author between the different types of papers makes a
% one-size-fits-all approach a daunting prospect. For instance, compsoc 
% journal papers have the author affiliations above the "Manuscript
% received ..."  text while in non-compsoc journals this is reversed. Sigh.

\author{Yuehai~Chen, \IEEEmembership{Member,~IEEE,}
        Qingzhong~Wang, \IEEEmembership{Member,~IEEE,}
        Jing Yang, \IEEEmembership{Member,~IEEE,}
         Badong Chen, \IEEEmembership{Senior Member,~IEEE,}   
         Haoyi Xiong, \IEEEmembership{Senior Member,~IEEE,}
         and Shaoyi~Du, \IEEEmembership{Member,~IEEE,}% <-this % stops a space
\IEEEcompsocitemizethanks{\IEEEcompsocthanksitem 
% M. Shell was with the Department
% of Electrical and Computer Engineering, Georgia Institute of Technology, Atlanta,
% GA, 30332.\protect\\
Yuehai~Chen and Jing Yang are with the School of Automation Science and Engineering, Xi’an Jiaotong University, Xi’an 710049, China.

E-mail: cyh0518@stu.xjtu.edu.cn, jasmine1976@xjtu.edu.cn
% note need leading \protect in front of \\ to get a newline within \thanks as
% \\ is fragile and will error, could use \hfil\break instead.
% E-mail: cyh0518@stu.xjtu.edu.cn
\IEEEcompsocthanksitem Badong Chen and Shaoyi~Du are with the Institute of Artificial Intelligence and Robotics, Xi’an Jiaotong University, Xi’an, Shanxi 710049, China.

E-mail: chenbd@mail.xjtu.edu.cn, dushaoyi@gmail.com
% <-this % stops a space
% \IEEEcompsocthanksitem Gang Hua is with Wormpex AI Research LLC, Bellevue, WA 98004 USA.

Qingzhong Wang and Haoyi Xiong are with Baidu Research, Beijing 100085, China.

E-mail: qingzwang@outlook.com, haoyi.xiong.fr@ieee.org

% E-mail: ganghua@gmail.com
\IEEEcompsocthanksitem Yuehai Chen and Qingzhong Wang contribute equally to this work.
\IEEEcompsocthanksitem Corresponding author: Jing Yang and Shaoyi Du.
}}
% \thanks{Manuscript received April 19, 2005; revised August 26, 2015.}}

% note the % following the last \IEEEmembership and also \thanks - 
% these prevent an unwanted space from occurring between the last author name
% and the end of the author line. i.e., if you had this:
% 
% \author{....lastname \thanks{...} \thanks{...} }
%                     ^------------^------------^----Do not want these spaces!
%
% a space would be appended to the last name and could cause every name on that
% line to be shifted left slightly. This is one of those "LaTeX things". For
% instance, "\textbf{A} \textbf{B}" will typeset as "A B" not "AB". To get
% "AB" then you have to do: "\textbf{A}\textbf{B}"
% \thanks is no different in this regard, so shield the last } of each \thanks
% that ends a line with a % and do not let a space in before the next \thanks.
% Spaces after \IEEEmembership other than the last one are OK (and needed) as
% you are supposed to have spaces between the names. For what it is worth,
% this is a minor point as most people would not even notice if the said evil
% space somehow managed to creep in.

% The paper headers
\markboth{Journal of \LaTeX\ Class Files,~Vol.~14, No.~8, August~2015}%
{Shell \MakeLowercase{\textit{et al.}}: Bare Advanced Demo of IEEEtran.cls for IEEE Computer Society Journals}
% The only time the second header will appear is for the odd numbered pages
% after the title page when using the twoside option.
% 
% *** Note that you probably will NOT want to include the author's ***
% *** name in the headers of peer review papers.                   ***
% You can use \ifCLASSOPTIONpeerreview for conditional compilation here if
% you desire.

% The publisher's ID mark at the bottom of the page is less important with
% Computer Society journal papers as those publications place the marks
% outside of the main text columns and, therefore, unlike regular IEEE
% journals, the available text space is not reduced by their presence.
% If you want to put a publisher's ID mark on the page you can do it like
% this:
%\IEEEpubid{0000--0000/00\$00.00~\copyright~2015 IEEE}
% or like this to get the Computer Society new two part style.
%\IEEEpubid{\makebox[\columnwidth]{\hfill 0000--0000/00/\$00.00~\copyright~2015 IEEE}%
%\hspace{\columnsep}\makebox[\columnwidth]{Published by the IEEE Computer Society\hfill}}
% Remember, if you use this you must call \IEEEpubidadjcol in the second
% column for its text to clear the IEEEpubid mark (Computer Society journal
% papers don't need this extra clearance.)

% use for special paper notices
%\IEEEspecialpapernotice{(Invited Paper)}

% for Computer Society papers, we must declare the abstract and index terms
% PRIOR to the title within the \IEEEtitleabstractindextext IEEEtran
% command as these need to go into the title area created by \maketitle.
% As a general rule, do not put math, special symbols or citations
% in the abstract or keywords.
\IEEEtitleabstractindextext{%
\begin{abstract}
Crowd counting models in highly congested areas confront two main challenges: weak localization ability and difficulty in differentiating between foreground and background, leading to inaccurate estimations. The reason is that objects in highly congested areas are normally small and high-level features extracted by convolutional neural networks are less discriminative to represent small objects. To address these problems, we propose a learning discriminative features framework for crowd counting, which is composed of a masked feature prediction module (MPM) and a supervised pixel-level contrastive learning module (CLM). The MPM randomly masks feature vectors in the feature map and then reconstructs them, allowing the model to learn about what is present in the masked regions and improving the model’s ability to localize objects in high-density regions. The CLM pulls targets close to each other and pushes them far away from background in the feature space, enabling the model to discriminate foreground objects from background. Additionally, the proposed modules can be beneficial in various computer vision tasks, such as crowd counting and object detection, where dense scenes or cluttered environments pose challenges to accurate localization. The proposed two modules are plug-and-play, incorporating the proposed modules into existing models can potentially boost their performance in these scenarios.
\end{abstract}

% Note that keywords are not normally used for peerreview papers.
}

% make the title area
\maketitle

% To allow for easy dual compilation without having to reenter the
% abstract/keywords data, the \IEEEtitleabstractindextext text will
% not be used in maketitle, but will appear (i.e., to be "transported")
% here as \IEEEdisplaynontitleabstractindextext when compsoc mode
% is not selected <OR> if conference mode is selected - because compsoc
% conference papers position the abstract like regular (non-compsoc)
% papers do!
\IEEEdisplaynontitleabstractindextext
% \IEEEdisplaynontitleabstractindextext has no effect when using
% compsoc under a non-conference mode.

% For peer review papers, you can put extra information on the cover
% page as needed:
% \ifCLASSOPTIONpeerreview
% \begin{center} \bfseries EDICS Category: 3-BBND \end{center}
% \fi
%
% For peerreview papers, this IEEEtran command inserts a page break and
% creates the second title. It will be ignored for other modes.
\IEEEpeerreviewmaketitle

\ifCLASSOPTIONcompsoc
\IEEEraisesectionheading{\section{Introduction}\label{sec:introduction}}
\else
\section{Introduction}
\label{sec:introduction}
\fi

\IEEEPARstart{R}{ecently}, crowd counting has been receiving increasing attention from researchers due to its wide-ranging applications \cite{TNNLS2}. Especially after the outbreak of the COVID-19 pandemic, accurately estimating the number of people in crowd scenes plays an important role in monitoring crowd gatherings and slowing down the spread of the disease \cite{TCSVT1, TIP2, TIP3, TIP4}. In recent years, CNN-based regression methods have made remarkable progress and established an undisputed dominance \cite{DM-Count, ADSCNet, SASNet}. However, they still face the problem of accuracy degradation in highly congested scenes.

\begin{figure}
    \centering    
    \includegraphics[width=1.0\linewidth]{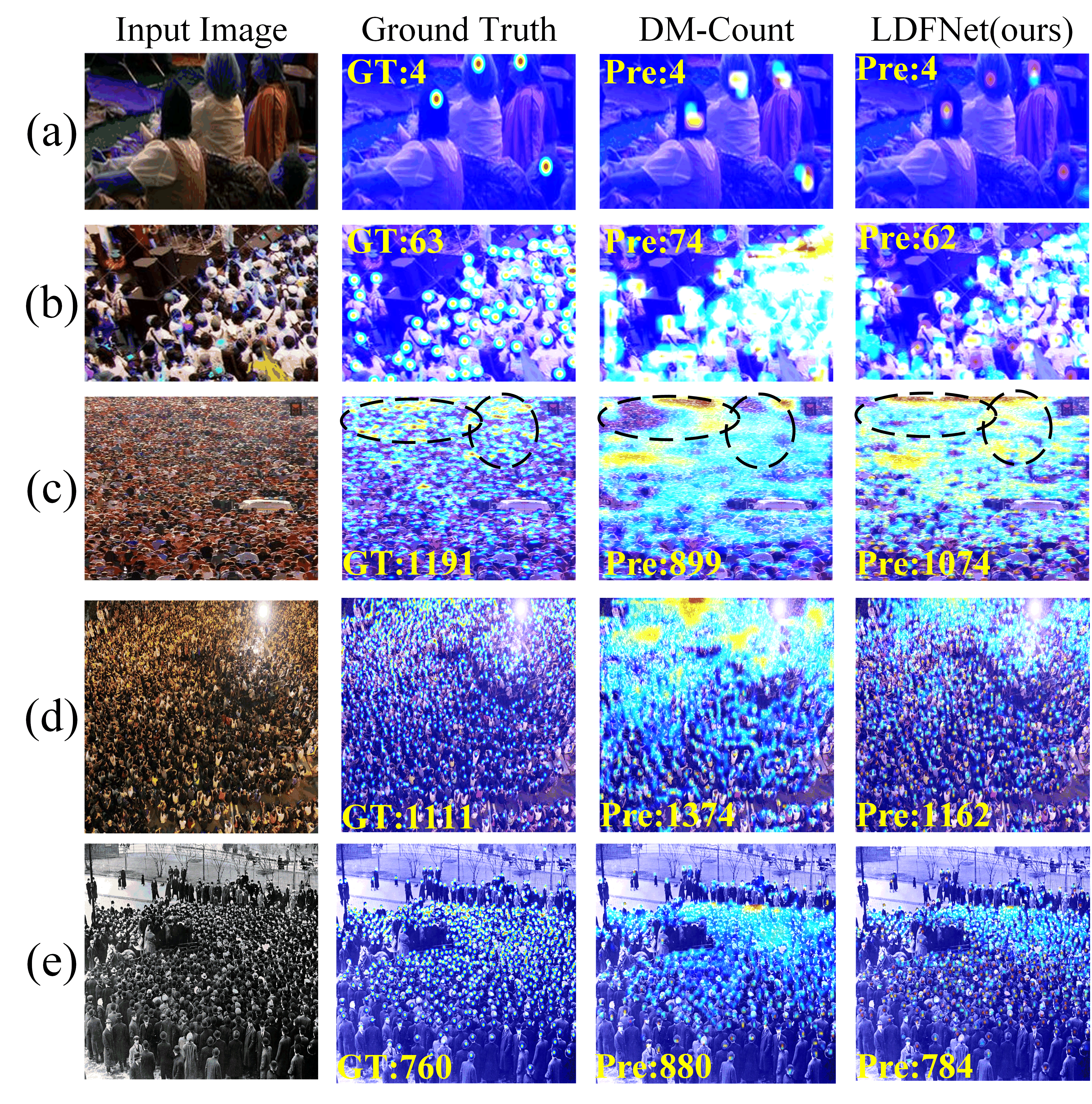}
    \caption{Problems of existing regression-based crowd counting models. From left to right: input images from ShanghaiTech Part A dataset \cite{MCNN}, Ground Truth density maps, the predicted density maps from DM-Count \cite{DM-Count}, and the predicted density maps from our LDFNet+DM-Count.}
    \label{fig:introduction}
\end{figure}

Figure \ref{fig:introduction} provides insights into the limitations of existing state-of-the-art crowd counting models. We observe that (1) DM-Count \cite{DM-Count} can accurately estimate the number of people in low-density regions, as shown in Fig. \ref{fig:introduction} (a), (2) for high-density regions, two main issues plague existing models. The first problem, illustrated in Fig. \ref{fig:introduction} (b), (d) and (e), is that it is difficult for these models to accurately localize each person in the crowded regions, and the predicted density maps often appear over-smooth. The second problem, demonstrated in Fig. \ref{fig:introduction} (c), is that existing models often treat the foreground of the overcrowded regions as background, leading to significant underestimation. We argue that the main reason for these issues is that the extracted features from the crowded regions are less discriminative, which results in the models being unable to localize small objects and distinguish foreground from background. Previous methods employ attention mechanisms \cite{ADCrowdNet, CTASNet, MAN} and multi-scale feature fusion \cite{SASNet, M-SFANet, fusion} to predict discriminative crowd features. For example, MAN \cite{MAN} employs multi-level attention mechanisms to predict discriminative crowd features. \cite{fusion} exploits the adaptive fusion of a large majority of encoded features to extract saliency information. These methods enhances counting performance, however, they still face the challenge that they cannot localize small objects in the high-cluttered environment and dense scenes.

The challenges posed by crowded scenes necessitate novel approaches for crowd counting that can improve prediction accuracy. To tackle these issues, we introduce a crowd counting framework called the learning discriminative features network (LDFNet) that includes feature enhancement modules. To improve the ability of localization in dense regions, we propose a masked feature prediction module (MPM) which randomly masks pixels in the feature map and then reconstructs them, enabling the model to understand the entire scene. The process of randomly masking feature vectors introduces an element of uncertainty, forcing the model to rely on the surrounding context to infer the missing information. This can encourage the model to capture more fine-grained details and learn to distinguish subtle differences between objects in crowded areas. Moreover, the reconstruction step allows the model to compare the original feature vectors with the reconstructed ones. By minimizing the reconstruction loss, the model can learn to generate accurate representations of the masked regions. This can help model know the content of the masked regions and contribute to improved localization capabilities.

% By reconstructing masked regions, MPM could know the content of the masked regions, thus improves the model's ability to localize small objects. 

\begin{figure}[t]
  \centering
  %\fbox{\rule{0pt}{2in} \rule{0.9\linewidth}{0pt}}
   \includegraphics[width=1.0\linewidth]{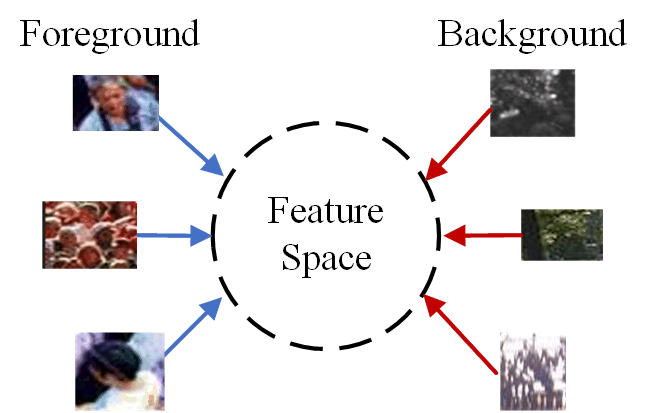}
   \caption{The proposed supervised pixel-level contrastive learning module pulls pull the targets closer to each other and push them further away from the background.}
   \label{fig:densecl}
\end{figure}

To address the issue of confusion between foreground and background in high-density regions, a supervised pixel-level contrastive learning module (CLM) is proposed to separate foreground and background in feature space. As shown in Fig.~\ref{fig:densecl}, CLM uses pixel-level contrastive learning to pull the targets closer to each other and push them further away from the background. The objective is to encourage the model to embed pixels from the same target close together while ensuring a clear separation between target and background pixels. By pulling targets close to each other in the feature space, the module can facilitate better clustering of similar objects in dense regions. This can aid in separating foreground and background information and mitigating confusion in high-density regions.
To sum up, in this paper, we address the problems of existing regression-based crowd counting models in high-density regions and the contributions are threefold.

\begin{itemize}
\item We propose a masked feature prediction module that randomly masks feature vectors in the feature map and then reconstructs them. This module allows the model to learn about what is present in the masked regions, thereby improving the model's ability to localize objects in high-density regions.
    
\item We propose a supervised pixel-level contrastive learning module that pulls targets close to each other and pushes them far away from background in the feature space. This module could separate foreground and background in feature space, thus mitigating the confusion in high-density regions.

\item The proposed modules can be beneficial in various computer vision tasks, such as crowd counting and object detection, where dense scenes or cluttered environments pose challenges to accurate localization. The proposed two modules are flexible, incorporating the proposed modules into existing models can potentially boost their performance in these scenarios.

\end{itemize}

\section{Related Works}

\noindent\textbf{CNN-based Crowd Counting.}
Lately, CNN-based regression methods have \wqz{achieved inspiring} performances \wqz{in crowd counting} \cite{MCNN, SANet, SASNet, SFCN, xuyi3, TIP5, TIP6, TIP7, TIP8, TIP9}. Different network architectures are designed to handle various challenges, such as scale changes and congestion. To deal with scale variation, a multi-column convolutional neural network (MCNN) \cite{MCNN} is proposed, which uses multiple branches to extract multi-scale features, while SASNet \cite{SASNet} uses a scale-adaptive selection network to automatically learn scale information. It is believed that contextual information benefits crowd counting, especially for congested regions \cite{ADMG}. In \cite{shang2016end}, a contextual pyramid CNN is used to predict both local and global counts. In \cite{CAN}, an end-to-end deep architecture that can adaptively capture both scale and contextual information is proposed, improving the performance on high-density regions. \cite{TAU} proposes a scale alignment method to alleviate the catastrophic sensitivity of crowd counters to scale shift. DDC \cite{DDC} treats density map generation as a denoising diffusion process for predicting accurate density maps. It \cite{DDC} adopts a diffusion model, which requires many steps of noise reduction and would take a relatively long training time.

\noindent\textbf{Transformer-based Crowd Counting.}
Transformer uses self-attention mechanisms and has indeed shown great potential in dealing with multi-scale change problems \cite{NIPS2017_3f5ee243}. MAN \cite{MAN} incorporates global attention from vanilla transformer, learnable local attention, and instance attention into a counting model. The attention region of the learnable local attention mechanism can change as crowd scale changes. AGCCM \cite{AGCCM} proposed a bi-transformer that distinguishes the advantages of multi-scale branches using global attention maps and recombines multi-scale predictions following a Wenn diagram. HMoDE \cite{HMODE} redesigns the multi-scale neural network by introducing a hierarchical mixture of density experts, which hierarchically merges multi-scale density maps for crowd counting. CrowdFormer \cite{CrowdFormer} proposes a vision Transformer-based encoder for extracting coarse and fine features. TransCrowd \cite{TransCrowd} reformulates the weakly-supervised crowd counting problem from the perspective of sequence-to-count based on Transformer.

These transformer-encoder-based methods leverage the self-attention mechanism to capture dependencies between all pairs of pixels in the feature map. This can help the model understand the relationship between different parts of a crowd and accurately estimate the crowd density in an image. However, these methods lack specialized extraction of detailed information, which helps localize small objects in the high-cluttered environment and dense scenes. Different from these transformer-encoder-based methods, the proposed MPM has been designed to balance the focus between context and detailed information. The proposed MPM randomly masks different parts of the feature map in different training instances. This randomness ensures that the Transformer encoder does not simply 'ignore' the masked regions but learns to adapt to the variability, enhancing its ability to capture detailed information. In addition, reconstructing masked features relies on understanding surrounding features, which illustrates the ability of the proposed MPM to capture contextual features.

% These transformer-based methods could solve the scale change problem well and achieve inspiring counting performances. However, these methods still face the challenge that they cannot localize small objects in the high-cluttered environment and dense scenes.

\noindent\textbf{Point-based Crowd Counting.}
Density maps are essentially intermediate representations that are constructed from an annotation dot map, whose optimal choice of bandwidth varies with the dataset and network architecture \cite{NoiseCC}. Thus, some point annotation directly based framework methods are proposed in crowd counting \cite{BL, DM-Count, UOT, GL}. The Bayesian loss (BL) uses a point-wise loss function between the ground-truth point annotations and the aggregated dot prediction generated from the predicted density map \cite{BL}. DM-count considers density maps and dot maps as probability distributions and uses balanced OT to match the shape of the two distributions \cite{DM-Count}. GL \cite{GL} and UOT \cite{UOT} adopt unbalanced OT to improve the performance of DM-Count \cite{DM-Count}. OT-M \cite{otm} propose the optimal transport minimization algorithm for crowd localization with density maps.

To achieve accurate individual localization in crowd counting, \cite{p2pnet} first proposes a purely point-based framework for joint crowd counting and individual localization. Following closely behind, \cite{CLTR}  introduce a KMO-based Hungarian matcher and porposes an elegant, end-to-end crowd localization transformer that solves the localization task in the regression-based paradigm. These two methods present desirable performance in distinguishing head centers in crowd scenes. However, they did not specifically design solutions for highly dense scenarios. The proposed MPM allows the model to learn about what is present in the masked regions with the masking and reconstructing structure. This could improve the model’s ability to localize objects in high density regions. Applying the proposed MPM into P2PNet localization method could improve its performances in dense scenes. The related results have been presented in Table \ref{localization} and \ref{localization_N}.

% \wqz{Though these models achieve satisfying performance \cite{RANet}, the problems of localization and misestimation in high-density regions trouble existing models.}

\noindent\textbf{Masked Autoencoders.}
Masked autoencoders are proposed for pre-training and have been widely applied in natural language processing and computer vision tasks \cite{bert,MAE,wei2022masked}. In BERT \cite{bert}, the tokens in the input sentences are randomly masked and then a transformer \cite{NIPS2017_3f5ee243} is employed to encode the masked sentences. Finally, the masked tokens are predicted using contextual information. Likewise, MAE \cite{MAE} uses a ViT \cite{ViT} to encode masked images and a small ViT as a decoder to reconstruct the masked image patches.  

Many methods follow the ideas of MAE and design various masked autoencoders, which have demonstrated excellent performance in many downstream tasks. SemMAE \cite{SemMAE} designs a masking strategy that varies from masking a portion of patches in each part to masking a portion of whole parts in an image. Such a design can gradually guide the network to learn various information which benefits image classification and semantic segmentation. MultiMAE \cite{MultiMAE} proposes multi-modal multi-task masked autoencoders including GRB, depth, and semantic segmentation information, achieving inspiring performances on classification, semantic, and dense regression tasks. MaskFeat \cite{wei2022masked} reconstruct the HoG features \cite{hogdetection} of the masked image patches. Following MAE, MASL \cite{MASL} randomly mask out spacetime patches in videos, which enables learning general visual representation performing well in image recognition. Audio-MAE \cite{AudioMAE} extends image-based masked autoencoders to self-supervised representation learning from autio spectrograms. This sets new state-of-the-art performance on six audio and speech classification tasks. MCMAE \cite{MCMAE} utilizes multi-scale hybrid convolution-transformer to learn more discriminative representations, gaining improved performances on object detection, semantic segmentation, and video understanding tasks. ConvNeXt \cite{ConvNeXt} proposes a fully convolutional masked autoencoder framework with a new Global Response Normalization layer, which can significantly improve the performance across object detection and semantic segmentation.

The main reason that limits the improvement of crowd counting accuracy is inaccurate estimation of dense areas \cite{CTASNet}. As argued in the introduction section, the extracted features from dense areas are less discriminative, which results in the models being unable to localize small objects in the dense areas. Masked Autoencoders benefits for discriminative features extraction. However, the counting task encompasses a multitude of scenarios, making the reconstruction of original image pixel values following MAE \cite{MAE} a challenging endeavor. We find that the feature map encapsulates valuable crowd semantic information, and precise feature representations can significantly improve the accuracy of crowd counting \cite{ADMG, CTASNet}. Consequently, we have specifically engineered a masking prediction module in the feature level, which randomly masks feature vectors and reconstruct them. If the model can successfully reconstruct a masked feature vector, it signifies that the model has comprehended the information represented by that vector. This module allows the model to learn about what is present in the masked regions, thereby improving the model’s ability to localize objects in high-density regions.

% To our best knowledge, this work is the first to integrate supervised masked autoencoders into crowd counting task. 
% Different from MAE \cite{MAE}, the proposed MPM applies the mask-prediction mechanism to CNN features instead of original images. The process of randomly masking feature vectors introduces an element of uncertainty, forcing the model to rely on the surrounding context to infer the missing information. This can encourage the model to capture more fine-grained details and learn about what is present in the masked regions, benefiting localization. In addition, using CNN features requires fewer computational resources, since the number of feature vectors is smaller than that of image patches.

% Whereas the proposed MPM applies the mask-prediction mechanism to CNN features instead of original images, which enables the extracted features to localize small objects in high-density regions. In addition, using CNN features requires fewer computational resources, since the number of feature vectors is smaller than that of image patches.}

\noindent\textbf{Contrastive Learning.}
Dating back to 2006, the idea of contrastive learning was proposed to learn invariant representations \cite{ hadsell2006dimensionality}. In \cite{dosovitskiy2014discriminative}, a model is proposed to learn to discriminate between a set of surrogate classes. By contrast, Z. Wu et al. \cite{wu2018unsupervised} propose an approach to learn instance-level discriminative features using contrastive learning. In recent self-supervised models \cite{chen2020simple,he2020momentum,xie2020pointcontrast}, a positive pair is usually generated using two augmented views of the same image, while a negative one is generated using different images. 

Prior studies have also incorporated surpervised contrastive learning into various domains, including object detection \cite{FSCE, DenseCL, SoCo}, semantic segmentation \cite{alonso2021semi, DenseCL, wang2021exploring}, and instance segmentation \cite{DenseCL}. FSCE \cite{FSCE} first integrates supervised contrastive learning into few-shot object detection. It proposes a simple yet effective approach, contrastive proposals encoding, to learn contrastive-aware object proposal encodings that facilitate the classification of detected objects. DenseCL \cite{DenseCL} proposes a dense contrastive learning, which performs dense pair-wise contrastive learning at the level of pixel. It demonstrates consistently superior performance when transferring to downstream dense prediction tasks including object detection, semantic segmentation and instance segmentation. \cite{SoCo} introduces a new contrastive learning method which maximizes the similarity of object-level features representing different augmentations of the same object. \cite{alonso2021semi} proposes a contrastive learning module that enforces the segmentation network to yield similar pixel-level feature representations for same-class samples across the whole dataset. \cite{wang2021exploring} proposes a supervised, pixel-wise contrastive learning method which could make full use of the global semantic similarities among labeled pixels. These tasks inherently have object/class level positive and negative samples/features, which suit for modeling the contrast.

As argued in the introduction section, existing models often treat the foreground of the overcrowded regions as background, leading to significant underestimation. As per the principles of crowd counting, the features of various heads should maintain consistency within the feature map, and they must be entirely distinct from the background. To facilitate this, we have specifically proposed a Supervised Pixel-Level Contrastive Learning Module (CLM). This module operates by pulling analogous targets closer, while concurrently pushing them further away from the background within the feature space. This significantly reduces the likelihood of the model misidentifying non-human elements as people, thereby enhancing the model's discrimination capabilities and overall counting performance.

% The CLM ensures robust generalization across different instances of the same class, in this case, the human head, regardless of occlusions and backgrounds. 

% Contrastive learning methods normally use global representations of images and are in the pre-train-fine-tune paradigm, while we design a supervised pixel-level contrastive learning module to discriminate targets from the background in high-density regions, mitigating the problem of misestimation.

\begin{figure*}[t]
\begin{center}
\includegraphics[width=0.95\linewidth]{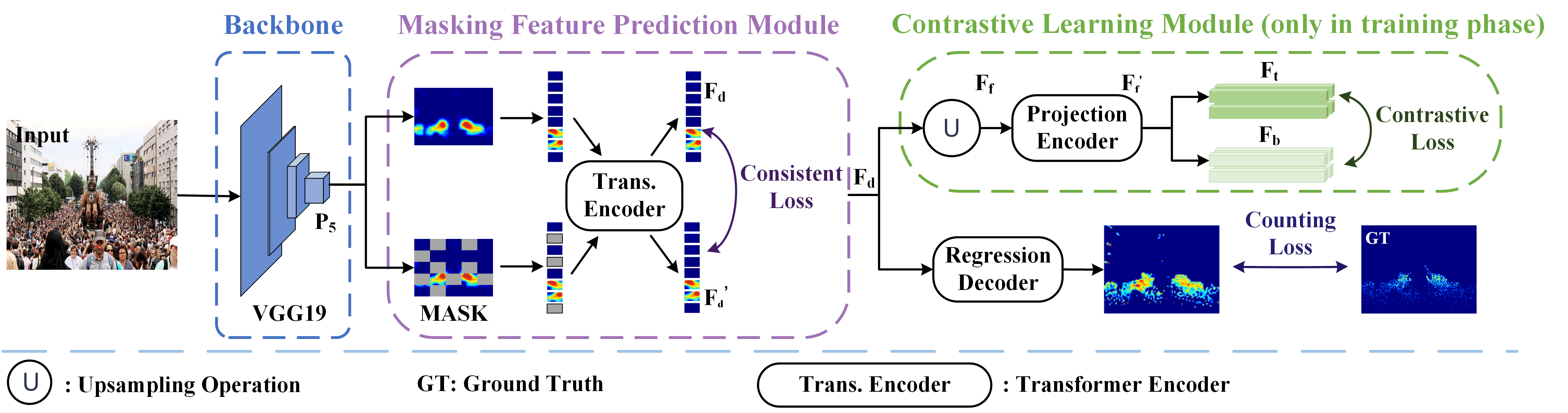}
\end{center}
   \caption{The proposed learning discriminative feature network (LDFNet) framework. The LDFNet mainly contains a backbone, a masked feature prediction module (MPM), and a supervised pixel-level contrastive learning module (CLM). All the components including the backbone, MPM, CLM, and regression decoder are trainable in the training phase. While in the inference phase, the CLM would be removed and other components are used for inference. The masking process is pixel-wise on feature map whose size is 1/32 of the input image. For intuitive visualization, we upsample the feature map to the size of the original image. As a result, this figure presents that the masking process seems patch-wise (32$\times$32).}
\label{fig:framework}
\end{figure*}

\section{Proposed Method} \label{sec:method}

Fig.~\ref{fig:framework} presents an overview of the proposed framework --- LDFNet, which consists of 3 components: (1) a backbone to extract image features, (2) MPM to improve the ability of localization, and (3) CLM to improve the discriminability of foreground and background.
\wqz{In terms of the backbone, similar to existing models \cite{BL, DM-Count}, we also use a vgg19 to extract image features and the proposed modules are performed on the extracted feature maps.} The size of the extracted features from the backbone network is typically much smaller than the size of the original image. This reduction in size significantly reduces the computational resources required to process the features and apply the proposed modules.

\subsection{Masked Feature Prediction Module}

As mentioned earlier, by randomly masking feature vectors in the feature map and then reconstructing them, the model is forced to fill in the missing information based on the surrounding context. Learning to reconstruct masked regions helps the model understand what is present in those regions, which in turn benefits localization. 

In high-density scenarios, objects may be occluded or densely packed together, making it challenging for the model to accurately identify individual objects. However, by training the model to reconstruct the masked regions, it becomes more adept at inferring the underlying information, even in these challenging situations. The enhanced understanding of the masked regions can aid in better object localization by allowing the model to reason about the presence and characteristics of objects that are partially occluded or closely neighboring other objects. This additional information helps the model differentiate between objects and improves its ability to precisely localize them in high-density regions. Building on this insight, we propose a masked feature prediction module (MPM). Similar to BERT \cite{bert} and MAE \cite{MAE}, we randomly mask certain feature vectors in the last feature map $P_5$ using a mask token $0$, and then use a transformer to encode the contextual information. Finally, we reconstruct the masked feature vectors. The entire process is illustrated in Fig.~\ref{fig:framework}.

\noindent \textbf{Masking and encoding.} 
In the training phase, we first flat the extracted feature map $P_5\in \mathbb{R}^{C\times \frac{W}{32}\times \frac{H}{32}}$ to  $P_5\in \mathbb{R}^{C\times N}$, where $N=\frac{W}{32}\times\frac{H}{32}$ represents the number of feature vectors, and randomly mask some feature vectors, yielding masked feature map $P_{5}^\prime\in \mathbb{R}^{C\times N}$. Then we employ a linear projection layer to obtain an embedding of each feature vector. Finally, the encoder with multiple self-attention layers and multi-layer perceptrons take the feature embedding as input to obtain the refined features $F_d^\prime$. Also, we feed the unmasked feature vectors into the encoder, yielding $F_d$. Randomly masking certain feature vectors create a situation where the model is deprived of direct information about those vectors. The model then needs to rely on the contextual cues and information from the surrounding feature vectors to make predictions about the masked vectors. This ability to infer the missing information based on context is a powerful aspect of the approach and can lead to improved localization.

%Since the masked tokens are easy to be integrated into a sequence, we adopt a Transformer \cite{ViT} as the encoder of MPM. To reduce the memory usage and computational cost, we employ a Transformer on feature $P_{5}$. To be specific, we reshape the spatial dimensions of the pooling feature $P_{5} \in \textbf{R}^{C \times \frac{W}{32} \times \frac{H}{32}}$ into one dimension, resulting in feature $ P_{5}^{'} \in \textbf{R}^{C \times \frac{WH}{1024}}$.\footnote [1]{$H$ and $W$ indicate the spatial height and width of an input image, $C$ is the channel number of feature map $P_{5}$.} 
%Then the flattened feature is embedded by a linear projection and processed in a series of Transformer blocks to obtain the discriminative features $F_{d}$. Finally, we concatenate the augmented features $F_{mc}$ and discriminative features $F_{d}$ to get feature $F_{f}$ which will be fed into a regression decoder for the final output.

\noindent \textbf{Reconstruction.}
When the model is trained to reconstruct the masked regions, it needs to capture the essential features and patterns necessary to accurately reconstruct the original information. This process encourages the model to learn contextual relationships and dependencies within the feature map. As a result, the model becomes more attentive to the details and fine-grained information in the masked regions. Improving the understanding of the masked regions can have a positive impact on localization in high-density regions. 
\wqz{Obviously, if the representations $F_d^\prime$ and $F_d$ are consistent with each other, then the model captures the contextual information and has the knowledge on the masked regions. To achieve this, we use Euclidean distance as the consistent loss function, which can be computed as:}
%If the encoder's outputs of the masked feature and unmasked feature are the same, this indicates that the encoder could predict the masked feature values by understanding the contextual information of different local regions. In this paper, Euclidean distance is adopted as the consistent loss function $\mathcal{L}_{mp}$ in the proposed MPM:
\begin{equation}
\mathcal{L}_{mp} =  \sum_{i\in\mathcal{M}}\left|F_{d}^{\prime i}-{F}_{d}^{i}\right|^{2}
\end{equation}

\noindent \wqz{where $\mathcal{M}$ denotes the indices of the masked feature vectors.}

The uncertainty introduced by MPM helps the model learn robust representations by forcing it to consider multiple plausible interpretations of the masked vectors based on the surrounding context. This encourages the model to capture higher-level semantic information and dependencies, enabling it to make accurate predictions about the missing features. Thus, MPM learns to know what is present in those regions and benefits localization.
\wqz{Note that, in training and inference, we use the output of the transformer with unmasked input for counting, while the consistent loss serves only as a regularizer. This ensures that our training and testing processes are consistent.} 
%\NOTE{Why do you just consider the consistency among masked feature vectors? Since you use a transformer, I think unmasked feature vectors are not consistent either. I think you can try to make all feature vectors consistent. Plus, what if you make $F_d^\prime$ consistent with $P_5$, does it work?}

\subsection{Supervised Pixel-level Contrastive Learning}
The aim of crowd counting is to accurately identify and count the number of human heads present in an image or a video \cite{DM-Count, Noise_tip}. However, crowd scenes can often present a certain level of noise due to occlusions and diverse backgrounds. It is important to distinguish human heads from diverse backgrounds. % As per the principles of crowd counting, the features of human heads should maintain consistency within the feature map, and they must be entirely distinct from the background.

Inspired by recent self-supervised models \cite{chen2020simple,he2020momentum} which apply contrastive learning to discriminate positive samples from negative ones, we propose a supervised pixel-level contrastive learning module (CLM) to distinguish human head regions (targets) from background. The aim of CLM is to pull the targets closer to each other and push them further away from the background. This separation can reduce confusion particularly in scenarios with high object density. As shown in Fig. \ref{fig:framework}, given an input image, the \cyh{upsampled} dense feature map \cyh{{$F_{f}\in \mathbb{R}^{C\times\frac{W}{8}\times\frac{H}{8}}$}} are extracted by the network and fed into a projection head. The projection head consists of two convolution layers \wqz{with kernel sizes $3\times 3$ and $1\times 1$ respectively} and a ReLU layer between them. \wqz{Finally, we obtain a dense feature map $F_f^\prime \in \mathbb{R}^{D\times\frac{H}{8} \times \frac{W}{8}}$.} % Per pixel in the a dense feature map $F_f^\prime$ represents an 8$\times$8 region in the original image.

% The proposed CLM pulls the targets closer to each other and push them further away from the background. This separation can reduce confusion particularly in scenarios with high object density. As shown in Fig. \ref{fig:framework}, given an input image, the \cyh{upsampled} dense feature map \cyh{{$F_{f}\in \mathbb{R}^{C\times\frac{W}{8}\times\frac{H}{8}}$}} are extracted by the network and fed into a projection head. The projection head consists of two convolution layers \wqz{with kernel sizes $3\times 3$ and $1\times 1$ respectively} and a ReLU layer between them. \wqz{Finally, we obtain a feature map $F_f^\prime \in \mathbb{R}^{D\times\frac{H}{8} \times \frac{W}{8}}$.}

As per the principles of crowd counting, the features of human heads should maintain similarity within the feature map, and they must be entirely distinct from the background \cite{ADCrowdNet}. To this aim, the proposed CLM is applied on the dense feature map $F_f^\prime$ whose size is $\frac{1}{8}$ of the original image. Per pixel in the dense feature map represents an 8$\times$8 region in the original image. If a feature vector in $F_f^\prime$ corresponds to the head region in the original image, it would be defined as a target representation and the corresponding label is 1 ($l_{xy}=1$); otherwise it would be defined as a background representation and the corresponding label is 0 ($l_{xy}=0$). Thus, the labeled pixels in the dense feature map represent a combination of regional context in the original image. The ultimate goal of CLM is to provide a rough estimation of the head location to help the model focus on the most crucial area. Containing the center of larger head regions still provides effective head location information \cite{Noise_tip}. This means it's capable of handling larger head regions effectively, even if they extend beyond the labeled center pixels. 
The target representations are treated as positive samples, and the contrastive loss for one positive sample is calculated as follows:

% Specifically, during training, we utilize the label $l_{xy}$ of each pixel in $F_f^\prime$, which indicates whether the pixel is a head region ($l_{xy}=1$) or background ($l_{xy}=0$), to apply pixel-level contrastive learning. Specifically, we define a feature vector in $F_f^\prime$ as a target representation if the corresponding label is 1; otherwise, we define it as a background representation. The target representations are treated as positive samples, and the contrastive loss for one positive sample is calculated as follows:
 
\begin{equation}
\mathcal{L}_{\mathrm{cl}}\left(i\right)= -\log \frac{ e^{\cos \left(\textbf{x}_i, \textbf{x}_p^g\right)}}{ e^{\cos \left(\textbf{x}_i, \textbf{x}_p^g\right)}+ e^{\cos \left(\textbf{x}_i, \textbf{x}_n^g\right)}},
\label{eq:dense_loss}
% \frac{1}{M}\sum_{i \in \Omega_p}
\end{equation}
\wqz{where $\textbf{x}_i$ denotes the $i$-th target representation, $cos(\cdot,\cdot)$ denotes the cosine similarity between two vectors, $\textbf{x}_p^g=\frac{1}{|\Omega_p|}\sum_{j\in \Omega_p} \textbf{x}_j$ is the global pooling of target representations, $\Omega_p$ denotes the set of target representations and $|\Omega_p|$ denotes the number of targets. Likewise, $\textbf{x}_n^g=\frac{1}{|\Omega_n|}\sum_{j\in \Omega_n} \textbf{x}_j$ is the global pooling of background representations. The aim of pooling operation which performs a statistical aggregation of these features, is to ensure that the obtained target and background features are more representative. Finally, the contrastive loss for an entire image is the average of the losses for all positive samples, \ie,}
\wqz{
\begin{equation}
   \mathcal{L}_{\mathrm{cl}}=\frac{1}{|\Omega_p|}\sum_{i\in \Omega_p}\mathcal{L}_{cl}(i).
\end{equation}
}
%\noindent where $\textbf{x}_p^g=F_g\left(\textbf{x}_i\right), i \in \Omega_p$; $\textbf{x}_n^g=F_g\left(\textbf{x}_i\right), i \in \Omega_n$; $\textbf{x}_p^g/\textbf{x}_n^g$ is the statistical features of targets and background, $ \Omega_p/\Omega_n $ is the collection of target/background positions. $ F_g $ is a global pooling operation, which is used to obtain the statistical features of targets and backgrounds. The contrastive learning loss $\mathcal{L}_{\mathrm{cl}}$ is averaged over all target pixels on the feature map.

As a result, the CLM can learn more discriminative features to distinguish targets from the background, thereby mitigating misestimations in high-density regions. It is worth noting that the proposed CLM only contains two convolution layers and a ReLU layer between them, which introduces only a few extra learnable parameters. Moreover, the loss functions can be efficiently implemented using matrix operations, resulting in negligible latency overhead. Additionally, we can drop the contrastive learning module during inference, thereby avoiding any extra computational complexity during inference.

\subsection{Loss Function}

%For optimizing the entire network, 
\wqz{In terms of the counting loss function}, we adopt the function in DM-Count \cite{DM-Count}:
% but modify the transport cost function as a Gaussian function following \cite{CTASNet}:

\begin{equation}
\begin{split}
\mathcal{L}_{d}\left(\boldsymbol{D}^{\prime}, \boldsymbol{D}\right)
= \left|\left\|\boldsymbol{D}^{\prime}\right\|_{1}-\|\boldsymbol{D}\|_{1}\right|+\lambda_{1} \mathcal{L}_{O T}\left(\boldsymbol{D}^{\prime}, \boldsymbol{D}\right) \\
+ \lambda_{2} \mathcal{L}_{T V}\left(\boldsymbol{D}^{\prime}, \boldsymbol{D}\right)
\label{OT}
\end{split}
\end{equation}
where $\boldsymbol{D}^{\prime}$ and $\boldsymbol{D}$ denote the predicted density map and dot-annotation map, respectively. $\|\cdot\|_{1}$ denote the $L_{1}$ norm. The first term in Eq. \ref{OT} is to minimize the difference in the total count between the prediction and ground truth. The second term (optimal transport loss) is used to minimize the distribution distance between $\boldsymbol{D}$ and $\boldsymbol{D}^{\prime}$. The third term (total variation loss) is able to increase the stability of the training procedure. % More details are presented in the supplemental.

The final loss function is given by the weighted sum of $\mathcal{L}_d$, $\mathcal{L}_{mp}$, and $\mathcal{L}_{cl}$, \ie,
\begin{equation}
\mathcal{L}=\mathcal{L}_{d}+\alpha \mathcal{L}_{mp}+\beta \mathcal{L}_{cl},
\label{combine}
\end{equation}
where $\alpha$ and $\beta$ are the weights for the consistent loss $\mathcal{L}_{mp}$ and contrastive learning loss $\mathcal{L}_{cl}$, respectively.

\section{Experiments}

In this section, we conduct experiments to evaluate the proposed LDFNet. We first present detailed experimental setups, including network architecture, training details, evaluate metrics, and \cyh{datasets}. Then, we compare the proposed method with recent state-of-the-art approaches. Finally, we conduct ablation studies to verify the effectiveness of the proposed modules.

\subsection{Experimental Setups}

\noindent \textbf{Network Architecture.}
For our proposed LDFNet, We adopt VGG-19 as our CNN backbone network which is pre-trained on ImageNet. We refer to \cite{NIPS2017_3f5ee243} for the structure of the Transformer encoder in our masked feature prediction module. \cyh{Specifically, we use a 4-layer Transformer with 512 hidden sizes and 2 attention heads.} \cyh{The intermediate size of feed-forward networks is 2048.} The regression decoder for the final output contains two $3 \times 3$ convolutional layers with 256 and 128 channels, and a $1 \times 1$ convolutional layer with 1 channel. Please note, the flexible nature of our proposed LDFNet allows for its incorporation into other baseline networks, including CSRNet \cite{CSRNet}, M-SFANet \cite{M-SFANet}, BL \cite{BL}, and MAN \cite{MAN}. In a specific application, we seamlessly integrated the LDFNet, comprising of the MPM and CLM, into the final encoding layer of these networks, with the remaining layers kept intact. The MAN \cite{MAN} network consists of four transformer encoder layers. To optimize computational efficiency, we substituted the final transformer encoder layer of the MAN network with our MPM, and appended the CLM, keeping the rest of the layers unaltered.

% Note that the proposed LDFNet is flexible, and we also incorporate proposed LDFNet into other baseline networks which are CSRNet \cite{CSRNet}, M-SFANet \cite{M-SFANet}, BL \cite{BL} and MAN \cite{MAN}. Specifically, we directly add the proposed LDFNet including MPM and CLM to the last encoding layer of these networks, leaving the others unchanged. MAN \cite{MAN} contains four transformer encoder layers. In order to reduce the amount of calculation, we replaced the last transformer encoder layer of MAN with our MPM and added the CLM, leaving the others unchanged.

\noindent \textbf{Training Details.}
Following \cite{DM-Count}, we first perform data augmentation using a random crop and limit the shorter side of each image within 2048 pixels. The random crop size is $512 \times 512$ in all datasets except ShanghaiTech A. As some images in ShanghaiTech A are in low %consist of a smaller 
resolution, the crop size for ShanghaiTech A changes to 224 $\times$ 224. In all experiments, we use the Adam \cite{kingma2014adam} algorithm with a learning rate $1e^{-5}$ to train our proposed model. %\NOTE{How many epochs and how to select a model for test?}

    \begin{table*}[t]
\centering
\caption{Results on the ShanghaiTech, UCF\_CC\_50, UCF-QNRF, and NWPU datasets. The best performance is indicated by bold numbers and the second best is indicated by underlined numbers.}\label{table:Comparisons} MAN* means the reproduced results that we use the official codes provided by MAN \cite{MAN} paper to get.
\centering
\setlength{\tabcolsep}{4mm}{\begin{tabular}{ccccccccccc}
 \hline 
 \multirow{2}{*} {Method} & \multicolumn{2}{c}{Part A} & \multicolumn{2}{c}{Part B} & \multicolumn{2}{c}{UCF\_CC\_50} & \multicolumn{2}{c}{UCF-QNRF} & \multicolumn{2}{c}{NWPU} \\
& MAE & RMSE & MAE & RMSE & MAE & RMSE & MAE & RMSE & MAE & RMSE \\
\hline MCNN\cite{MCNN} & $110.2$ & $173.2$ & $26.4$ & $41.3$ & $377.6$ & $509.1$ & $277.0$ & $426.0$ & $232.5$ & $714.6$ \\
Switch-CNN\cite{switch-CNN} & $90.4$ & $135.0$ & $21.6$ & $33.4$ & $318.1$ & $439.2$ & 228 & 445 & $-$ & $-$ \\
CSRNet\cite{CSRNet} & $68.2$ & $115.0$ & $10.6$ & $16.0$ & $266.1$ & $397.5$ & $120.3$ & $208.5$ & $121.3$ & $387.8$\\
SANet\cite{SANet} & $67.0$ & $104.2$ & $8.4$ & $13.6$ & $258.4$ & $334.9$ & $-$ & $-$& $-$ & $-$ \\
CAN\cite{CAN} & $62.3$ & 100 & $7.8$ & $12.2$ & $212.2$ & $243.7$ & 107 & 183 & $106.3$ & $386.5$ \\
BL\cite{BL} & $62.8$ & $101.8$ & $7.7$ & $12.7$ & $229.3$ & $308.2$ & $88.7$ & $154.8$  & $105.4$ & $454.0$\\
SFCN\cite{SFCN} & $67.0$ & $104.5$ & $8.4$ & $13.6$  &  $258.4$ & $334.9$ & $102.0$ & $171.4$ & $105.7$ & $424.1$\\
% ADSCNet\cite{ADCrowdNet} & $55.4$ & $97.7$ & $6.4$ & $11.3$  & $198.4$ & $267.3$ & $\textbf{71.3}$ & $\textbf{132.5}$ & $-$ & $-$ \\
% CG-DRCN-CC-Res101\cite{JHU2} & $60.2$ & $94.0$ & $7.5$ & $12.1$  & $-$ & $-$ & $95.5$ & $164.3$ & $-$ & $-$ \\
% SASNet\cite{SASNet} & $\underline{53.6}$ & $\underline{88.4}$ & $\textbf{6.4}$ & $\textbf{10.0}$  & $\underline{161.4}$ & $\underline{234.5}$ & $85.2$ & $147.3$ & $-$ & $-$\\
M-SFANet\cite{M-SFANet} & $59.7$ & $95.7$ & $6.8$ & $11.9$ & $162.3$ & $276.8$ & $85.6$ & $151.2$ & $-$ & $-$ \\
NoiseCC\cite{NoiseCC}& $61.9$ & $99.6$ & $7.4$ & $11.3$  & $-$ & $-$ & $85.8$ & $150.6$ & $96.9$ & $534.2$\\
DM-Count\cite{DM-Count}& $59.7$ & $95.7$ & $7.4$ & $11.8$  & $211.0$ & $291.5$ & $85.6$ & $148.3$ & $88.4$ & $388.4$\\
% S3\cite{S3} & $57.0$ & $96.0$ & $\textbf{6.3}$ & $10.6$  & $-$ & $-$ & $80.6$ & $139.8$ & $83.5$ & $346.9$\\
% UOT\cite{UOT}& $58.1$ & $95.9$ & ${6.5}$ & ${10.2}$  & $-$ & $-$ & $83.3$ & $142.3$ & $83.5$ & $346.9$\\
GL\cite{GL}& $61.3$ & $95.4$ & $7.3$ & $11.7$  & $-$ & $-$ & $84.3$ & $147.5$ & $79.3$ & $346.1$\\
ChfL \cite{ChfL}& $57.5$ & $94.3$ & $6.9$ & $11.0$  & $-$ & $-$ & ${80.3}$ & ${137.6}$ & ${76.8}$ & ${343.0}$\\
GauNet \cite{GauNet}& ${54.8}$ & ${89.1}$ & $\textbf{6.2}$ & $\underline{9.9}$  & ${186.3}$ & ${256.5}$ &$81.6$ & $153.7$ & $-$ & $-$\\
% 
% MAN \cite{MAN} & $56.8$ & $90.3$ & $-$ & $-$  & $-$ & $-$ & $\textbf{77.3}$ & $\textbf{131.5}$ & $\textbf{76.5}$ & $\textbf{323.0}$ \\
MAN* \cite{MAN} & $56.2$ & $89.9$ & $-$ & $-$  & $-$ & $-$ & $78.0$ & ${138.0}$ & ${78.1}$ & ${337.1}$ \\
AGCCM \cite{AGCCM}& $61.4$ & $97.5$ & $7.2$ & $11.8$  & $194.7$ & $246.8$ & $-$ & $-$ & $-$ & $-$\\
HMoDE \cite{HMODE}& $54.4$ & $87.7$ & $\textbf{6.2}$ & $\textbf{9.8}$  & $159.6$ & $\underline{211.2}$ & $-$ & $-$ & $\underline{73.4}$ & $331.8$\\
DDC \cite{DDC}& $52.9$ & $85.6$ & $-$ & $-$  & $\underline{157.1}$ & ${220.6}$ & $\textbf{65.8}$ & $\textbf{126.5}$ & $-$ & $-$\\
\hline 
LDFNet+CSRNet & $ {58.5}$ & $ {94.3}$ & ${7.5}$ & $12.9$ & $ {195.7}$ & $ {278.5}$ & $ {95.1}$ & $ {175.3}$ & $ {94.6}$ & $ {429.7}$\\
LDFNet+BL & $ {56.8}$ & $ {88.5}$ & ${7.2}$ & $12.2$ & $ {189.5}$ & $ {267.3}$ & $ {80.3}$ & $ {135.9}$ & $ {77.5}$ & $ {348.1}$\\
LDFNet+M-SFANet & $\underline{52.7}$ & $\underline{82.9}$ & ${6.7}$ & $11.5$ & ${164.9}$ & ${234.8}$ & ${77.8}$ & ${134.7}$ & $ {78.4}$ & $ {342.1}$\\
LDFNet+DM-Count & ${54.3}$ & ${84.1}$ & $\underline{6.3}$ & ${10.5}$ & ${182.9}$ & ${243.5}$ & ${79.2}$ & ${137.2}$ & ${76.9}$ & $\underline{328.8}$\\
LDFNet+MAN & $\textbf{51.3}$ & $\textbf{80.7}$ & $6.8$ & $11.6$ & $\textbf{150.2}$ & $\textbf{210.6}$ & $\underline{73.7}$ & $\underline{132.6}$ & $\textbf{72.1}$ & $\textbf{314.4}$\\
\hline 
\end{tabular}}
\end{table*}

\noindent \textbf{Evaluation Metrics.}
The widely used mean absolute error (MAE) and the root mean squared error (RMSE) are adopted to evaluate the performance. The MAE and RMSE are defined as follows: 
\begin{equation}
    M A E=\frac{1}{N} \sum_{i=1}^{N}\left|\boldsymbol{C}^{\prime}_{i}-\boldsymbol{C}_{i}\right|
    \end{equation}
    
    \begin{equation}
    R M S E=\sqrt{\frac{1}{N} \sum_{i=1}^{N}\left|\boldsymbol{C}^{\prime}_{i}-\boldsymbol{C}_{i}\right|^{2}}
\end{equation}
where $N$ is the number of test images, $\boldsymbol{C}^{\prime}_{i}$ and $\boldsymbol{C}_{i}$ are the estimated density map count and the ground truth annotation respectively.

% \begin{equation}
% M A E=\frac{1}{N} \sum_{i=1}^{N}\left|\boldsymbol{C}^{\prime}_{i}-\boldsymbol{C}_{i}\right|
% \end{equation}

% \begin{equation}
% R M S E=\sqrt{\frac{1}{N} \sum_{i=1}^{N}\left|\boldsymbol{C}^{\prime}_{i}-\boldsymbol{C}_{i}\right|^{2}}
% \end{equation}
% where $N$ is the number of test images, $\boldsymbol{C}^{\prime}_{i}$ and $\boldsymbol{C}_{i}$ are the estimated density map%count 
% and the ground truth annotation respectively.

\noindent
\textbf{Dataset.} We conduct experiments on several widely used crowd counting datasets, including ShanghaiTech A \cite{MCNN}, ShanghaiTech B \cite{MCNN}, UCF\_CC\_50 \cite{UCF-CC-50}, UCF-QNRF \cite{UCF-QNRF} and NWPU \cite{NWPU}. ShanghaiTech A \cite{MCNN} dataset is collected from the Internet and consists of 482 (300 for train, 182 for test) images with highly congested scenes. The images in ShanghaiTech A are highly dense with crowd counts between 33 to 3139. The ShanghaiTech B \cite{MCNN} dataset contains 716 (400 for train, 316 for test) images taken from busy streets in Shanghai. The images in ShanghaiTech B are less dense with the number of people varying from 9 to 578. UCF\_CC\_50 is an extremely dense crowd dataset but includes 50 images of different resolutions \cite{UCF-CC-50}. The numbers of annotations range from 94 to 4,543 with an average number of 1,280. Due to the limited training samples, 5-fold cross-validation is performed following the standard setting in \cite{UCF-CC-50}. UCF-QNRF is a challenging dataset which has a much wider range of counts than currently available crowd datasets \cite{UCF-QNRF}. Specifically, the dataset contains 1,535 (1,201 for train, 334 for test) jpeg images whose number ranges from 816 to 12,865. The NWPU dataset is the largest-scale and most challenging crowd counting dataset publicly available \cite{NWPU}. The dataset consists of 5,109 (3,109 for train, 500 for val , 1,500 for test) images whose number ranges from 0 to 20,003. Note that the ground truth of test images set are not released and researchers could submit their results online for evaluation. JHU-Crowd++ \cite{JHU2} contains 4,372 images with 1.51 million dot annotations. Images in JHU-Crowd++ are divided into different levels based on the number of people: low-density images (0-50 people), medium-density images (51-500 people), and high-density images (500+ people).

\subsection{Comparisons with State-of-the-Arts}

    \begin{table*}[t]
\caption{\wqz{The performance on JHU-Crowd++ dataset \cite{JHU2}. We report the performances on three levels of density: low, medium, and high.}}\label{table:Comparisons-jhu}
\centering
\begin{tabular}{ccccccccc}
 \hline 
 \multirow{2}{*} {Method} & \multicolumn{2}{c}{Low} & \multicolumn{2}{c}{Medium} & \multicolumn{2}{c}{High} & \multicolumn{2}{c}{Overall}\\
& MAE & RMSE & MAE & RMSE & MAE & RMSE & MAE & RMSE\\ 
\hline
DM-Count &9.6 &24.4 & 35.8 &57.7 &293.6 &691.9 &67.4 &270.8\\
DM-Count+LDFNet &10.3 &30.4 &31.3 &51.9 &259.9 &623.1 &56.3 &223.1\\
Improvement &-7.3\% &-24.6\% &\textbf{+12.6\%} &\textbf{+10.1\%} &\textbf{+11.5\%} &\textbf{+9.9\%} &\textbf{+16.5\%} &\textbf{+13.6\%} \\
P2PNet &16.4 &50.5 & 36.2 &57.1 &315.4 &722.3 &70.2 &282.7\\
P2PNet+LDFNet &18.2 &54.9 & 30.4 &49.8 &267.4 &654.2 &60.2 &252.6\\
Improvement &-9.9\% &-8.7\% &\textbf{+16.0\%} &\textbf{+12.8\%} &\textbf{+15.2\%} &\textbf{+9.4\%} &\textbf{+14.2\%} &\textbf{+10.6\%} \\
MAN &10.2 &26.5 & 30.2 &44.7 &253.1 &588.6 &52.7 &223.2\\
MAN+LDFNet &11.6 &29.8 & 25.4 &37.4 &237.9 &559.2 &49.1 &210.8\\
Improvement &-13.7\% &-12.5\% &\textbf{+15.9\%} &\textbf{+16.3\%} &\textbf{+6.0\%} &\textbf{+5.0\%} &\textbf{+6.8\%} &\textbf{+5.6\%} \\
\hline
\end{tabular}
% \vspace{-1em}
\end{table*}

Table~\ref{table:Comparisons} reports the experiment results. By incorporating the LDFNet into CSRNet \cite{CSRNet}, BL \cite{BL}, M-SFANet \cite{M-SFANet}, DM-Count \cite{DM-Count} and MAN \cite{MAN}, our methods achieve best counting performance in most dense crowd datasets (Part A, UCF\_CC\_50 and NWPU). Specifically, on the largest-scale and most challenging crowd counting dataset NWPU \cite{NWPU}, our ``LDFNet+MAN" achieves the best performance with 1.77\% MAE and 5.24\% MSE improvement compared with the state-of-the-art approach, HMoDE \cite{HMODE}. While for dense UCF-QNRF dataset, DDC \cite{DDC} achieves the best performance. The reason is that DDC \cite{DDC} uses conditional diffusion models which model complex distributions well to predict density maps. However, DDC adopts a diffusion model, which requires many steps of noise reduction and would take a relatively long inference time. By incorporating the LDFNet into MAN \cite{MAN}, our ``LDFNet+MAN'' achieves the second best performance on UCF-QNRF dataset.

On the sparse dataset Part B, our ``LDFNet+DM-Count" achieves a comparable performance with GauNet \cite{GauNet}. The reason is that Part B is composed of sparse scenes and the problems of localization and misestimation are not significant. To further verify this, we further conduct experiments on JHU-Crowd++ \cite{JHU2} with 3 levels of density and the performance is shown in Table \ref{table:Comparisons-jhu}. We can see that the proposed ``LDFNet+DM-Count" performs remarkably better than the baseline (DM-Count) on medium- and high-density images, but is inferior to DM-Count on low-density images. 
Note that the proposed LDFNet is specificlly designed for high-cluttered environment and dense scenes. As a result, the proposed LDFNet is more inclined to improve counting accuracy in dense areas, which benefits the overall counting performances. Although employing LDFNet makes RMSE in TABLE \ref{table:Comparisons-jhu} low density worsens for 24.6\%, the absolute increase at low densities is only 6.0. Compared with the absolute 68.8 reduction in RMSE at high density, the absolute increase at low densities is small. We also present the results on the overall dataset in Table \ref{table:Comparisons-jhu}. One could be observed that employing LDFNet make MAE and RMSE in TABLE \ref{table:Comparisons-jhu} overall better for 16.5\% and 13.6\%, respectively. Therefore, we can make the conclusion that to some extent, the proposed LDFNet addresses the problem of low counting accuracy in dense regions in existing regression-based crowd counting models. We also have added LDFNet to other methods, e.g., MAN and P2PNet in Table \ref{table:Comparisons-jhu}. The results shown in Table \ref{table:Comparisons-jhu} present the same performance trend under different densities.

We could find that our method consistently improves all baseline
methods. It is proved that our method can be an effective plug-in to the existing methods. Specifically, the proposed ``LDFNet+DM-Count" outperforms the baseline model --- DM-Count \cite{DM-Count} on all datasets, \eg, on Part B, LDFNet achieves 6.3/10.5 MAE/RMSE, while DM-Count obtains 7.4/11.8. Interestingly, looking at the performances on Part A, UCF\_CC\_50, and UCF-QNRF which are composed of high-density scenes, our ``LDFNet+DM-Count" is remarkably superior to DM-Count, \eg, ``LDFNet+DM-Count" achieves 54.3/84.1 MAE/RMSE on Part A, improving the baseline performance by 9.0\%/12.2\% and `LDFNet+DM-Count" also boosts the performance on UCF-QNRF by 7.5\%/7.5\%, while the improvement on the smaller and more challenging dataset --- UCF\_CC\_50 is by 13.3\%/16.5\%. We can conclude that the proposed model is more appropriate for high-density scenes since the problems addressed by LDFNet are more common in high-density regions.

\subsection{Ablation Study}
We conduct ablation studies on UCF-QNRF dataset to verify the effectiveness of the proposed modules.

\subsubsection{The Effectiveness of MPM and CLM}
We conduct ablation studies on the designed modules --- MPM and CLM to figure out the effectiveness of each module. We evaluate the performance of different models by plugging in MPM and CLM on the UCF-QNRF dataset. Table \ref{tab:CLM} presents the results of the models with and without the proposed modules. We observe that using MPM or CLM can significantly improve the performance of the four baseline models. MPM is particularly effective in reducing MAE/RMSE, for instance, plugging MPM into CSRNet reduces MAE/RMSE from 120.3/208.5 to 100.6/184.5, while using CLM achieves 107.3/193.7. This could be because MPM introduces more learnable parameters into the crowd counting model. Using both MPM and CLM further boosts the counting accuracy, e.g., plugging MPM and CLM into M-SFANet reduces MAE/RMSE from 85.6/151.2 to 77.8/134.7. Therefore, we can conclude that the proposed MPM and CLM are flexible and can be applied to existing models, significantly outperforming the baselines.

\begin{table}[t]
\caption{The Effectiveness of MPM and CLM}
\centering
\scalebox{0.9}{
\begin{tabular}{ccccc}
\hline \multirow{2}*{\text { Networks }} & Without & With MPM & With CLM & With both \\
 &MAE/RMSE & MAE/RMSE & MAE/RMSE & MAE/RMSE\\
\hline 
CSRNet \cite{CSRNet} & $120.3$/$208.5$ & $100.6$/$184.5$ & $107.3$/$193.7$ & $95.1$/$175.3$\\
BL \cite{BL} & $88.8$/$154.8$ & $82.9$/$143.6$ & $83.2$/$146.1$ & $80.3$/$140.2$\\
M-SFANet \cite{M-SFANet} & $85.6$/$151.2$ & $81.4$/$140.6$ & $83.1$/$142.2$ & $77.8$/$134.7$\\
DM-Count \cite{DM-Count} & $85.6$/$147.5$ & $81.3$/$141.8$& $81.7$/$142.6$ & $79.2$/$137.2$\\
\hline
\end{tabular}}
\label{tab:CLM}
\end{table}

\subsubsection{The Localization Ability of MPM} 
We incorporate MPM into four \wqzz{existing crowd counting frameworks}: BL \cite{BL}, DM-Count \cite{DM-Count}, P2PNet \cite{p2pnet} and GL \cite{GL}. MPM is \wqzz{applied} to the last layer of the backbone network in BL, DM-Count and GL. \wqzz{For P2PNet, we plug MPM into the regression head which is used for localization.} To evaluate the localization performance, we follow the approach presented in \cite{GL}, which is based on the local maximum in the predicted density map. We use Precision, Recall and F1-measure as evaluation metrics. 

We conduct experiments on two widely used datasets: UCF-QNRF and NWPU-Crowd and compare our approach with various advanced models. The comparisons are shown in Table \ref{localization} and Table \ref{localization_N}. \wqzz{Obviously, the proposed MPM is able to improve the localization ability, which boosts both precision and recall to a new Pareto front. \Eg, compared with GL \cite{GL} which achieves 0.782/0.748 precision/recall, introducing MPM into GL achieves 0.793/0.760 on UCF-QNRF. Additionally, compared with the models like CrossNet-HR \cite{CrossNet} and P2PNet \cite{p2pnet} which aim to localize objects for crowd counting, using MPM obtains notable performance, \eg, 0.742 vs. 0.712 vs. 0.739 F1-measure obtained by P2PNet+MPM, P2PNet and CrossNet-HR.}

\wqzz{In a word, the proposed MPM benefits localization and can be applied to any existing framework. Furthermore, we can also make the conclusion in Fig. \ref{fig:vis}.}
%We evaluated the localization performance on the NWPU dataset by adding the proposed MPM to the last network layer of GL \cite{GL} and the regression head for localization of P2PNet \cite{p2pnet}. To evaluate the localization, we used precision, recall, and F-measure as metrics. The results are shown in Table \ref{localization_N}, where we can see that incorporating MPM into each of the mentioned methods resulted in a significant improvement in all localization evaluation metrics. Furthermore, ``P2PNet+MPM" achieved the best overall performance as measured by the F-measure.
% The results suggest that the MPM is beneficial in improving the localization performance of the networks. Furthermore, the visual evidence provided in Fig. \ref{fig:vis} supports the idea that the MPM can enhance the localization ability of the networks.

\begin{table}[t]
\caption{Localization performance on UCF-QNRF dataset.}
\begin{center}
\scalebox{0.9}{\begin{tabular}{llll}
\hline Method & Precision & Recall & F1-measure \\
\hline MCNN \cite{MCNN} & $0.599$ & $0.617$ & $0.591$ \\
% ResNet \cite{resnet} & $0.616$ & $0.669$ & $0.612$ \\
DenseNet \cite{densenet} & $0.702$ & $0.581$ & $0.637$ \\
% Encoder-Decoder \cite{encoder} & $0.718$ & $0.630$ & $0.670$ \\
CL \cite{CL} & $0.758$ & $0.598$ & $0.668$ \\
VGG19+L2 & $0.605$ & ${0.670}$ & $0.636$ \\
\hline
BL \cite{BL} & ${0.767}$ & $0.654$ & ${0.706}$ \\
BL+MPM(ours) & $\textbf{0.781}$ \textcolor{red}{(+1.83\%)}& $\textbf{0.672}$ \textcolor{red}{(+2.75\%)}& $\textbf{0.722}$ \textcolor{red}{(+2.27\%)}\\
\hline
DM-Count \cite{DM-Count} & $0.731$ & $0.638$ & $0.682$ \\
DM-Count+MPM(ours) & $\textbf{0.756}$ \textcolor{red}{(+3.41\%)}& $\textbf{0.667}$ \textcolor{red}{(+4.55\%)}& $\textbf{0.700}$ \textcolor{red}{(+2.64\%)}\\
\hline
GL \cite{GL} & ${0.782}$ & ${0.748}$ & ${0.760}$ \\
GL+MPM(ours) & $\textbf{0.793}$ \textcolor{red}{(+1.40\%)}& $\textbf{0.760}$ \textcolor{red}{(+1.60\%)}& $\textbf{0.776}$ \textcolor{red}{(+2.11\%)}\\
\hline
% LDFNet(ours) & $\textbf{0.805}$ & $\textbf{0.767}$ & $\textbf{0.783}$ \\

\end{tabular}}
\label{localization}
\end{center}
\end{table}

\begin{table}[t]
\caption{Localization performance on NWPU-Crowd dataset.}
\begin{center}
\scalebox{0.9}{\begin{tabular}{llll}
\hline Method & Precision & Recall & F1-measure \\
\hline Faster RCNN \cite{Faster} & ${0 . 9 5 8}$ & $0.035$ & $0.068$ \\
TinyFace \cite{tiny} & $0.529$ & ${0 . 6 1 1}$ & $0.567$ \\
% VGG+GPR & $0.558$ & $0.496$ & $0.525$ \\
RAZNet \cite{RANet} & $0.666$ & $0.543$ & ${0.599}$ \\
D2CNet \cite{D2CNet} & $0.729$ & $0.662$ & $0.700$ \\
TopoCount \cite{topocount} & $0.695$ & $0.687$ & $0.691$ \\
CrossNet-HR \cite{CrossNet} & $0.748$ & $0.757$ & $0.739$ \\
\hline
GL \cite{GL} & ${0.800}$ & ${0.562}$ & ${0.660}$ \\
GL+MPM(ours) & $\textbf{{0.816}}$ \textcolor{red}{(+2.00\%)}& $\textbf{{0.588}}$ \textcolor{red}{(+4.62\%)}& $\textbf{{0.683}}$ \textcolor{red}{(+3.48\%)}\\
\hline
P2PNet \cite{p2pnet} & $0.729$ & $0.695$ & $0.712$ \\
P2PNet+MPM(ours) & $\textbf{0.764}$ \textcolor{red}{(+4.80\%)}& $\textbf{0.721}$ \textcolor{red}{(+3.74\%)}& $\textbf{0.742}$ \textcolor{red}{(+4.21\%)}\\
\hline
% LDFNet(ours) & $\textbf{0.825}$ & $\textbf{0.613}$ & $\textbf{0.695}$ \\
% \hline
\end{tabular}}
\label{localization_N}
\end{center}
\end{table}

\subsubsection{Hyper-parameter Study in MPM}
These hyper-parameter studies in MPM are conducted with ``DM-Count+LDFNet'' on Shanghai Part A and UCF-QNRF dataset.

\begin{figure}[t]
\centering
% \fbox{\rule{0pt}{2in} \rule{0.9\linewidth}{0pt}}
\includegraphics[width=\linewidth]{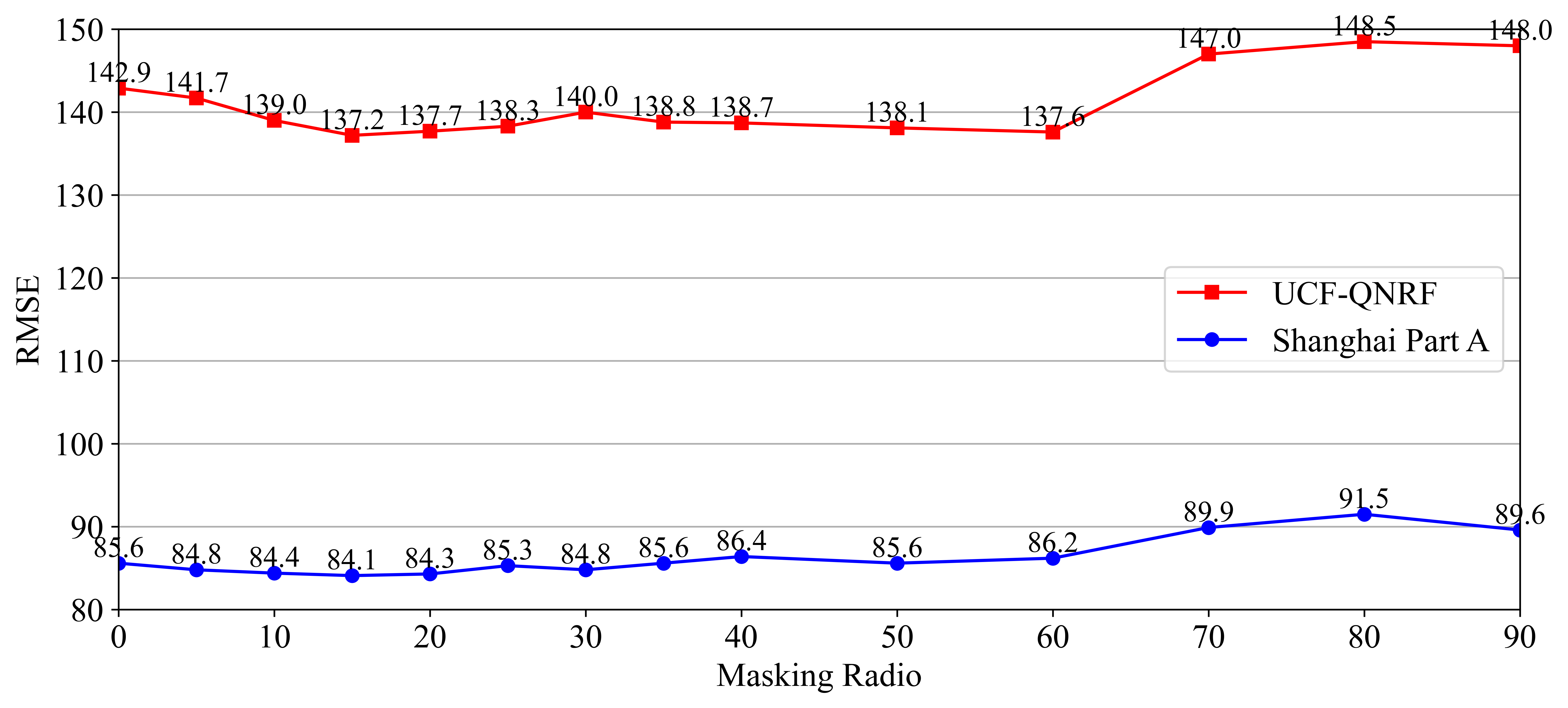}
% \vspace{-1em}
\caption{The performances obtained by using different masking ratios. %A low masking ratio (15\%) achieves the the best performance. %lowest RMSE performance.
}
\label{fig:mask}
% \vspace{-1em}
\end{figure}

\noindent \textbf{Masking Ratio.} We conducted experiments to evaluate the effect of different masking ratios in MPM. Figure \ref{fig:mask} presents the results, where a masking ratio of 0\% means that we only employ the transformer encoder without masked input and consistent loss. The results show that using a masking ratio ranging from 10\% to 60\% achieves better performance than using 0\% and 70\%+. However, a high masking ratio results in worse performance, as it is difficult to reconstruct the masked feature vectors with only limited information. We also found that the optimal masking ratio for our MPM is 15\%. It is worth noting that the optimal masking ratio for MPM is different from that of MAE \cite{MAE}. In MAE, image patches are masked for reconstruction, and each patch only contains a small amount of semantic information, making it easy to reconstruct the masked patches. However, our MPM performs a mask-prediction mechanism on a deep feature map, which contains more semantic information. Thus, it is difficult to reconstruct the masked deep representations with limited contextual information.
Moreover, the feature map represents the crowd semantic information which is similar across different crowd counting datasets. As a result, the masking ration chosen on UCF-QNRF is suitable for other scenes. To further verify the choice of masking ratio, we have conducted experiments on Shanhhai Part A dataset. The results presented in Figure 4 also show that the optimal masking ratio for our MPM is 15\%.

\begin{figure}
    \centering
    \includegraphics[width=0.95\linewidth]{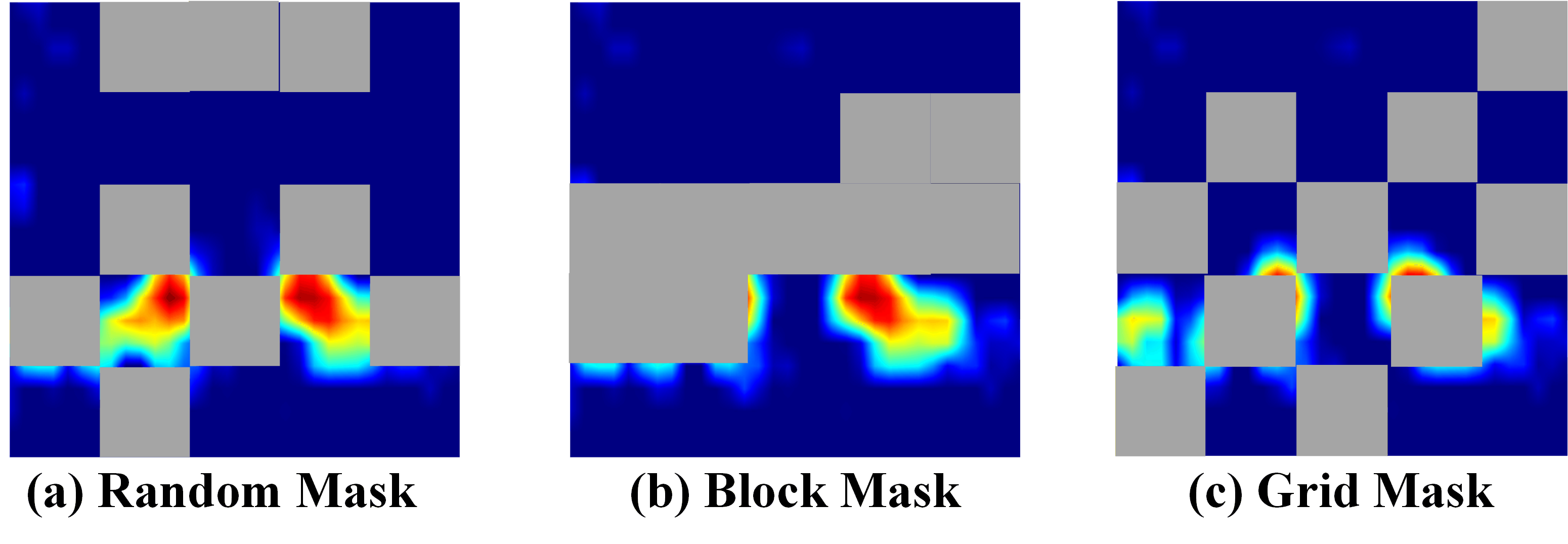}
    \caption{Different masking strategies. The masking process is pixel-wise on feature map whose size is 1/32 of the input image. For intuitive visualization, we upsample the feature map to the size of the original image. As a result, this figure presents that the masking process seems patch-wise (32$\times$32).}
    \label{fig:masking_strategies}
\end{figure}

\begin{figure*}[t]
    \centering
    \includegraphics[width=\linewidth]{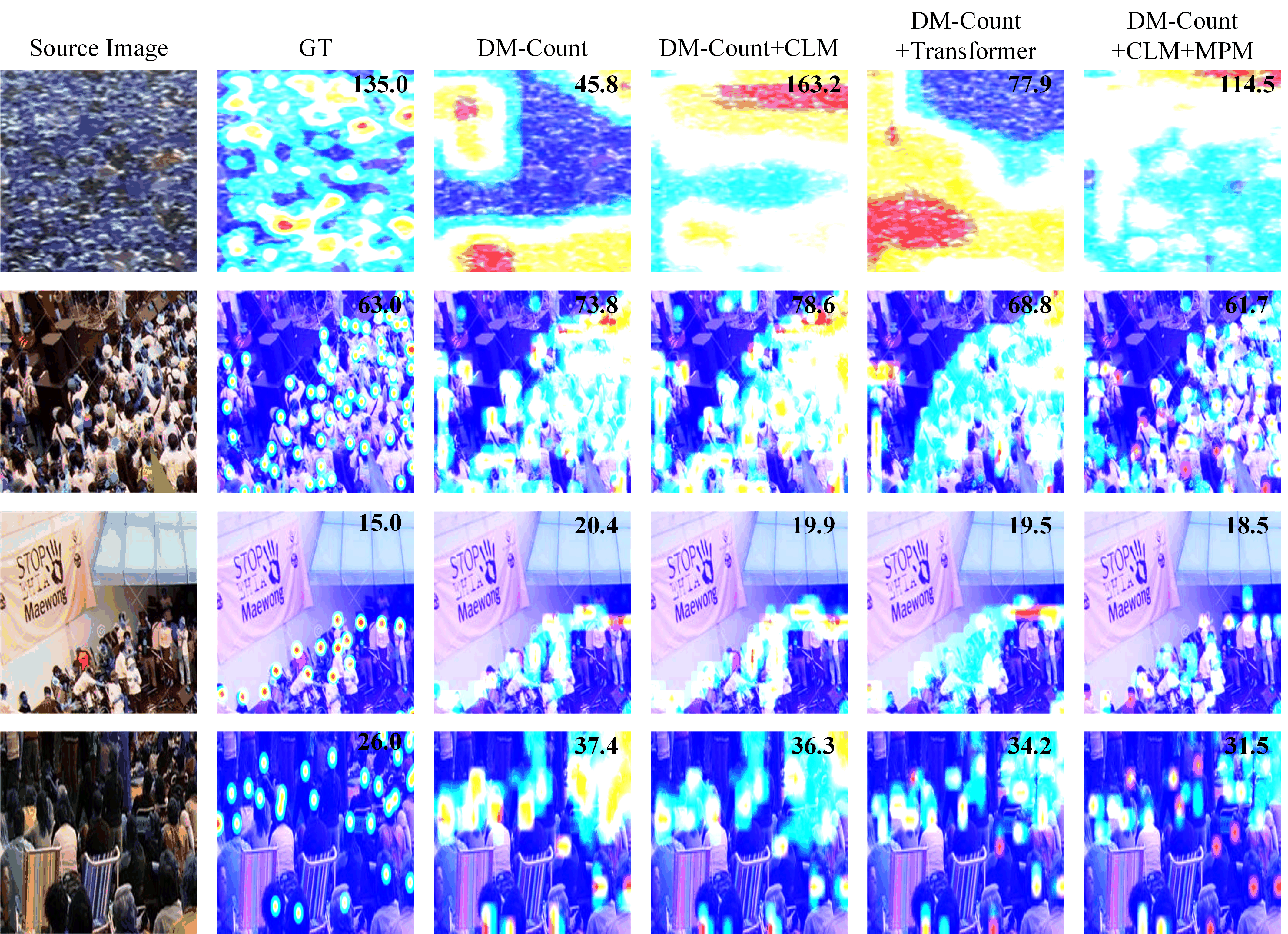}
    \caption{Visualization of the predicted results. DM-Count + Transformer means that we only plug the transformer into the baseline model and do not use the mask-prediction mechanism and contrastive loss.}
    \label{fig:vis}
\end{figure*}

\noindent \textbf{Masking strategy.}
We also tested different masking strategies shown in Figure~\ref{fig:masking_strategies} to evaluate their influence on the MPM performance. Table \ref{tab:strategy} shows the results of using different masking strategies. It can be observed that the simple random masking strategy achieves the best performance, indicating that it is the most effective in our MPM. Therefore, we adopt the random masking strategy in our experiments.

\begin{table}[htbp]
    \centering
        \caption{Results of using different masking strategies.}
        {
       \setlength{\tabcolsep}{7mm}{\begin{tabular}{ccc}
        \hline
        Masking strategy & MAE & RMSE \\
        \hline
        Random & $\textbf{79.2}$ & $\textbf{137.2}$ \\
        Block & $81.1$ & $138.3$ \\
        Grid & $81.5$ & $138.9$ \\
        \hline
        \end{tabular}
        }}
    
    \label{tab:strategy}
\end{table}

\noindent \textbf{Consistent loss.}
\wqz{In MPM, there are three ways to compute consistent loss: (1) only using the masked feature vectors, (2) using all feature vectors, (3) similar to MAE \cite{MAE}, reconstructing the mask input features $P_5$. Table~\ref{tab:mask} presents the performance using the three approaches to compute consistent loss. We can see that only using masked feature vectors to compute consistent loss outperforms using all feature vectors. In addition, considering all feature vectors in consistent loss results in slightly higher computational complexity. In contrast, reconstructing the masked input features $P_5$ also achieves competitive performance but is slightly worse than using masked vectors.}

\begin{table}[htbp]
    \centering
    \caption{Results of using different approaches to compute consistent loss.}
    \setlength{\tabcolsep}{7mm}{\begin{tabular}{ccc}
        \hline
        approach & MAE & RMSE \\
        \hline
        masked vectors & $\textbf{79.2}$ & $\textbf{137.2}$ \\
        all vectors & $79.9$ & $138.2$ \\
        $P_5$ & $79.4$ & $137.6$ \\
        \hline
        \end{tabular}}
    
    \label{tab:mask}
\end{table}

\noindent \textbf{The Number of Transformer Encoder Layers.}
To investigate the impact of the number of transformer encoder layers, we conduct experiments and present the results in Table \ref{tab:number}. We note that 0 corresponds to the case where we do not use MPM in LDFNet. As expected, using MPM significantly improves the model's performance compared to the case without MPM, as evidenced by the MAE/RMSE values of 79.2/137.2 and 83.7/145.8, respectively. We also observe that the optimal number of encoder layers is 4. Moreover, increasing the number of layers beyond 4 leads to a decrease in performance and a higher computational cost.

\begin{table}[htbp]
    \centering
    \caption{Results of using different numbers of encoder layers.}
    \setlength{\tabcolsep}{7mm}{\begin{tabular}{ccc}
        \hline
        \# layers & MAE & RMSE \\
        \hline
        0 & $83.7$ & $145.8$\\
        2 & $79.8$ & $138.2$\\
        4 & $\textbf{79.2}$ & $\textbf{137.2}$ \\
        6 & $80.1$ & $138.3$\\
        8 & $81.5$ & $140.1$ \\
        \hline
        \end{tabular}}
    
    \label{tab:number}
\end{table}

\begin{table}[ht]
\centering
\caption{Ablation study on contrastive learning loss.}
\setlength{\tabcolsep}{4mm}{\begin{tabular}{ccc}
\hline
Methods & MAE & RMSE \\
\hline
DM-Count \cite{DM-Count}  & $85.6$ & $148.3$ \\
+single-image global features & ${81.7}$ & $\textbf{142.6}$ \\
+single-image local features &  $82.0$ & $142.9$\\
+cross-image global features & $83.1$ & $145.7$ \\
+cross-image local features & $85.3$ & $146.1$ \\
+cross-image global features collection & $\textbf{81.6}$ & $142.9$ \\
\hline
\end{tabular}}

\label{tab:CLM1}
\end{table}

% \begin{figure*}[t]
% \begin{center}
% % \fbox{\rule{0pt}{2in} \rule{0.9\linewidth}{0pt}}
% \includegraphics[width=\linewidth]{vis1.png}
%    \caption{Visualization of the predicted results. DM-Count + Transformer means that we only plug the transformer into the baseline model and do not use the mask-prediction mechanism and contrastive loss.}
%    \end{center}
% \label{fig:vis}
% % \vspace{-1em}
% \end{figure*}

\subsubsection{Ablation Study on CLM} In this part, we explored the design of target and background representation, and the effect of dilation operation on CLM.

\noindent \textbf{Design of target and background representation.} 
Table \ref{tab:CLM1} displays the performance of various methods used to compute contrastive loss. As shown in Equation \ref{eq:dense_loss}, we use the global pooling of background representations as the negative sample and the global pooling of target representations as the positive sample. This process is referred to as the "single-image global features" in Table \ref{tab:CLM1}. 
As an alternative, we can calculate the contrastive loss for each pair of feature vectors within the same image. In other words, if two feature vectors in a feature map represent targets, they are considered positive to each other; otherwise, they are negative. This method is labeled as "single-image local features" in Table \ref{tab:CLM1}. 
Following the model of SimCLR cite{chen2020simple}, we can also employ target and background representations from other images to compute the contrastive loss. This approach results in the creation of "cross-image global features" and "cross-image local features". 
In addition, we calculate the average of global pooling of target representations within a training batch. Given the high variability in the backgrounds of different images, we use this "cross-image global features collection" as the positive sample, and the global pooling of background representations in each image as the negative sample. This method is dedicated to improving the "cross-image global features collection".

Our observations indicate that the application of Contrastive Learning Methods (CLM) significantly enhances performance, changing the metrics from 85.6/148.3 to 81.7/142.6. We note that global statistical features within a single image outperform local features, with performance metrics of 81.7/142.6 versus 82.0/142.9, respectively. This can be attributed to the fact that global statistical features possess more representative information, making them more effective in distinguishing targets from the background. Furthermore, the process of utilizing local feature vectors to compute contrastive loss is associated with increased computational complexity.
Interestingly, the "cross-image global features collection" method delivers performance on par with the "single-image global features", with performance metrics of 81.6/142.9 versus 81.7/142.6, respectively. This could be because global statistical features within a training batch also pack representative information. These findings underscore the importance of selecting representative features for contrastive learning.

\noindent \textbf{Effect of dilation operation on CLM.} 
To further verify whether the dilation operation is beneficial to CLM, we have conducted an ablation experiment on UCF-QNRF dataset. The detailed results have been presented in Table \ref{tab:CLM_dilation}.

\begin{figure}[htbp]
    \centering    
    \includegraphics[width=1.0\linewidth]{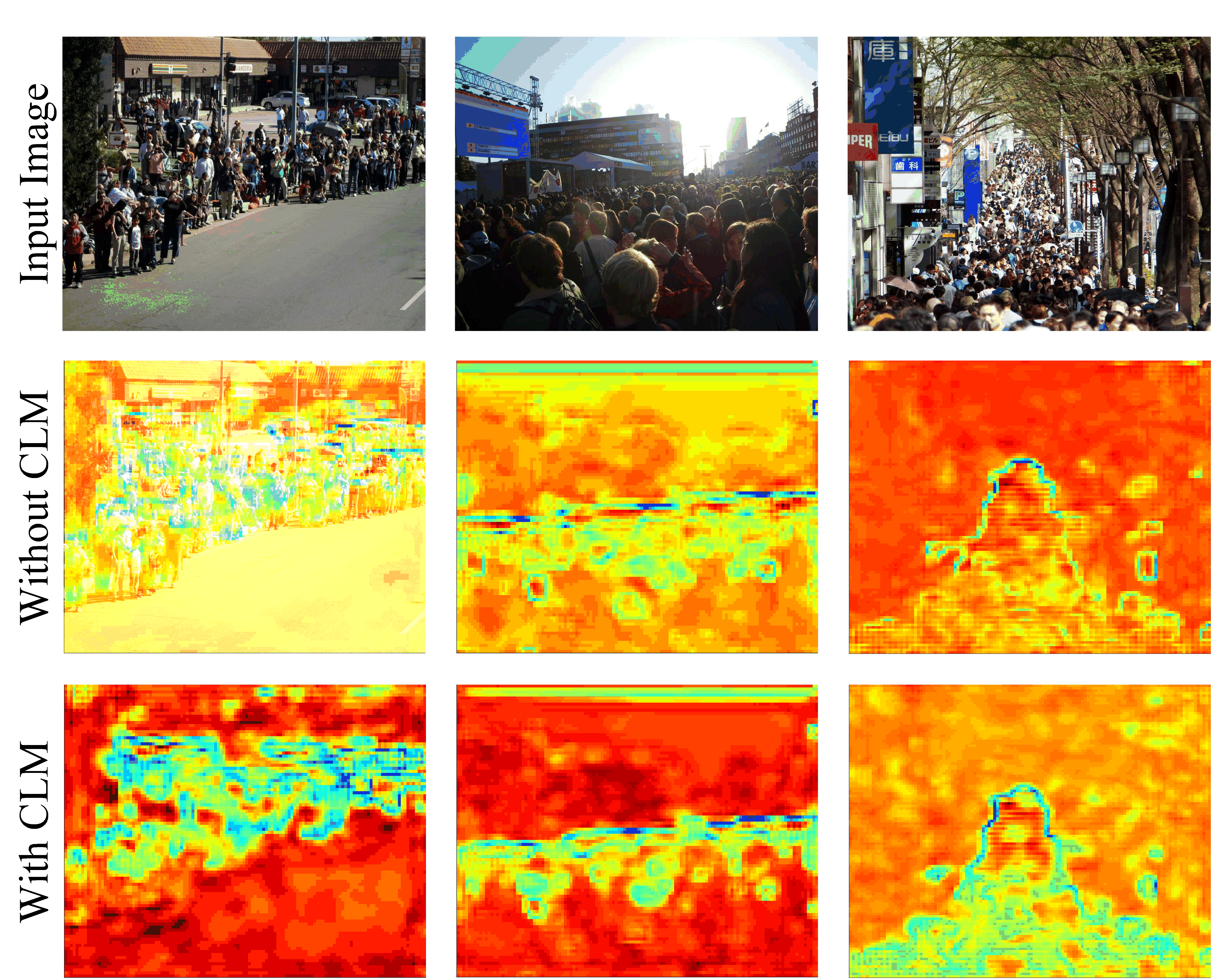}
    \caption{The visualization results of the separated representations from CLM.}
    \label{fig:CLM}
    \end{figure}

We initially decided to remove the CLM from the LDFNet for examination purposes. A comparative analysis between the original configuration, designated ''LDFNet with CLM (dilation = 1)'', and the alternate ''LDFNet without CLM'', reveals a superior performance by the former. This comparison implies that the introduction of the proposed CLM enhances counting accuracy. A plausible explanation for this could be the competence of the CLM in distinguishing between the human head and background within the feature space, thereby minimizing confusion in regions of high-density. We have also supplied a graphical illustration of the feature representations ($F_d$) with and without the CLM, which is depicted in Figure \ref{fig:CLM}. Specifically, a $1 \times 1$ convolution kernel is employed to decrease the channel dimensions of the feature ($F_d$) down to 1. The dimensionally reduced feature map is upsampled to original image size and showcased in Figure \ref{fig:CLM}.
We noticed that with the inclusion of CLM, the ''target'' feature response primarily concentrates on human head regions, as opposed to encompassing the entire human body. It is also evident that without the CLM, the feature representations of the head and background are somewhat similar. However, with the addition of CLM, the feature contrast between the head and the background becomes more pronounced. This observation confirms that our proposed CLM can effectively differentiate the features of the head and background. 

To validate the effect of the dilation operation in the context of CLM, we incorporated the dilation operation at various ratios into the original CLM. The term ``LDFNet with CLM (dilation = k)'' implies that the labels in the $k \times k$ region surrounding the labeled center pixel have been assigned a value of 1. The descriptor ``dilation = adaptive'' indicates the use of the adaptive kernel algorithm \cite{M-SFANet} to determine the size of each head region on the feature map, with the labels of these regions subsequently being set to 1. Table \ref{tab:CLM_dilation} displays that ``LDFNet with CLM (dilation = 3)'' and ``LDFNet with CLM (dilation = 5)'' lead to a deterioration in counting performance. This could be due to the fact that a dilation operation with a high dilation ratio identifies some background areas as head regions. ``LDFNet with CLM (dilation = adaptive)'' delivers performances commensurate with ``LDFNet with CLM (dilation = 1)''. However, the ``adaptive dilation'' approach incurs additional computational overhead.

% Specifically, we first remove the CLM from the LDFNet. We could observe that original setting, ``LDFNet with CLM (dilation = 1)'', performs better than ``LDFNet without CLM''. This indicates that the proposed CLM benefits counting accuracy. The reason may be that CLM could separate foreground and background in feature space, thus mitigating the confusion in high-density regions. We also present some visualization results of the separated representations in Figure 7. 

% To verify whether the dilation operation is beneficial to CLM, we employ dilation operation with different ratios into original CLM. ``LDFNet with CLM (dilation = k)'' means that the labels in the $k\times k$ region around labeled center pixel have been set as 1. ``dilation = adaptive'' means that we use the adaptive kernel algorithm \cite{M-SFANet} to calculate the size of each head region on the feature map, and then set the labels of these region to 1. The results in Table \ref{tab:CLM_dilation} show that ``LDFNet with CLM (dilation = 3)'' and ``LDFNet with CLM (dilation = 5)'' worsen the counting performance. The reason may be that dilation operation with high dilation ratio treats some background as head regions. ``'LDFNet with CLM (dilation = adaptive)'' achieves comparable performances with ``LDFNet with CLM (dilation = 1)''. However, the ``adaptive dilation'' would introduce additional computational overhead. 

\begin{table}[ht]
\setlength\tabcolsep{4pt}
\centering
\caption{Effect of dilation operation on CLM on UCF-QNRF dataset.}
{\begin{tabular}{ccc}
\hline
Method & MAE & RMSE \\
\hline
LDFNet without CLM  & $81.3$ & $141.8$ \\
LDFNet with CLM (dilation = 1) & $\textbf{79.2}$ & $\textbf{137.2}$ \\
LDFNet with CLM (dilation = 3) &  $82.6$ & $143.9$\\
LDFNet with CLM (dilation = 5) & $84.5$ & $148.6$ \\
LDFNet with CLM (dilation = adaptive) & $79.6$ & $137.4$ \\
\hline
\end{tabular}}
\label{tab:CLM_dilation}
\end{table}

\subsubsection{The Values of $\alpha$ and $\beta$}

We conducted experiments with the proposed ``DM-Count+LDFNet'' on the UCF-QNRF dataset to determine the optimal hyper-parameters $\alpha$ and $\beta$ in the combined loss functions. The detailed results are shown in Table~\ref{tab:p1} and \ref{tab:p2}. Firstly, we fixed $\beta$ to 0.01 and varied $\alpha$ from 0.01, 0.1 to 1. The resulting MAE varied from 81.4, 79.2 to 81.8. As $\alpha=0.1$ achieved the best result, we fixed $\alpha$ to 0.1 and tuned $\beta$ from 0.001, 0.01, 0.1 to 1. The MAE varied from 82.0, 79.2, 79.9 to 80.3. Therefore, we set $\alpha= 0.1$ and $\beta=0.01$, and used them for all the datasets.

\begin{table}[htbp]
    \centering
    \caption{Results of using different $\alpha$.}
    \setlength{\tabcolsep}{7mm}{\begin{tabular}{ccc}
        \hline
        $\alpha$ & MAE & RMSE \\
        \hline
        0.01 & $81.4$ & $141.2$\\
        0.1 & $\textbf{79.2}$ & $\textbf{137.2}$\\
        1 & $81.8$ & $142.1$ \\
        \hline
        \end{tabular}}
    
    \label{tab:p1}
\end{table}

\begin{table}[htbp]
    \centering
    \caption{Results of using different $\beta$.}
    \setlength{\tabcolsep}{7mm}{\begin{tabular}{ccc}
        \hline
        $\beta$ & MAE & RMSE \\
        \hline
        0.001 & $82.0$ & $143.7$\\
        0.01 & $\textbf{79.2}$ & $\textbf{137.2}$\\
        0.1 & $79.9$ & $139.8$\\
        1 & $80.3$ & $140.7$ \\
        \hline
        \end{tabular}}
    
    \label{tab:p2}
\end{table}

\subsection{Visualization}
Recall that the motivation of this paper is to address the problems of localization and confusion on foreground and background in crowd counting, to intuitively show the effectiveness of the proposed LDFNet in addressing the problems, we visualize the predicted density maps of high-density regions in Fig.~\ref{fig:vis}. In the first row, we can see that using CLM notably alleviates the problem of confusion on foreground and background. While looking at the 2nd, 3rd, and 4th rows, we can find that MPM benefits localization, and this ability is from mask-prediction mechanism instead of transformers. Though only using a transformer to refine the CNN features can improve the performance, the predicted density maps are over-smooth. In contrast, introducing mask-prediction mechanism results in more accurate localization. Overall, the proposed LDFNet can address the mentioned problems of localization and confusion on foreground and background, leading to more accurate counting results.

% \hspace*{\fill} \\ 
% \noindent \textbf{Complexity analysis.}
% To evaluate the complexity of our method, we have conducted ablation study on ShanghaiTech PartA dataset in Table \ref{tab6}. To exclude interference from other factors, we conducted the experiment on the same experimental environment, and reported the results in PartA benchmark \cite{MCNN}. The parameters and FLOPs are computes with the input size of 512×512 on a single NVIDIA 3090 GPU. The inference time is the average time of 100 runs on testing 1024×768 image sample.

% As shown in Table \ref{tab6}, our model does not have an advantage in model parameters and inference speed. However, our model has achieved better performance in the crowd counting. Moreover, our model could also achieve real-time crowd counting at a speed of 0.048 seconds per picture. It does not affect the application of our method in reality.

\begin{table*}
\centering
\caption{Flexibility of MPM and CLM. We add MPM and CLM into different detection frameworks on pascal-voc 07 \cite{voc07} dataset.}
\setlength{\tabcolsep}{0.7mm}{\begin{tabular}{c|c|c|cccccccccccccccccccccc} 
method & data & mAP & areo & bike & bird & boat & bottle & bus & car & cat & chair & cow & table & dog & horse & mbike & person & plant & sheep & sofa & train & tv \\
\hline \hline 
Faster R-CNN \cite{Faster} & 07 & 72.5 & 77.0 & 81.2 & 73.5 & 56.5 & 56.6 & 77.4 & 86.1 & 81.6 & 55.1 & 75.5 & 65.0 & 80.1 & 82.6 & 80.6 & 82.8 & 48.8 & 71.4 & 69.0 & 79.8 & 70.0 \\
Faster R-CNN+Transformer  & 07 & 72.8 & 78.6 & 79.7 & 73.1 & 57.4 & 55.0 & 75.9 & 86.5 & 82.5 & 56.0 & 77.1 & 65.1 & 82.2 & 83.9 & 77.4 & 79.3 & 49.7 & 73.3 & 69.4 & 82.2 & 70.5 \\
Faster R-CNN+CLM & 07 & 73.5 & 81.1 & 82.2 & 72.3 & 59.1 & 61.4 & 79.4 & 84.2 & 81.3 & 58.9 & 75.6 & 65.3 & 77.4 & 82.5 & 79.9 & 83.6 & 49.9 & 71.6 & 72.2 & 80.3 & 71.3 \\
Faster R-CNN+MPM & 07 & 74.2 & 77.9 & 82.7 & 73.7 & 61.1 & 60.5 & 79.3 & 87.0 & 82.9 & 57.8 & 76.5 & 68.4 & 79.2 & 84.6 & 77.5 & 83.9 & 52.3 & 70.7 & 73.0 & 81.1 & 73.5 \\
Faster R-CNN+CLM+MPM & 07 & 75.1 & 82.5 & 83.7 & 74.8 & 63.8 & 62.7 & 79.8 & 86.7 & 81.4 & 60.2 & 76.1 & 67.4 & 80.2 & 83.3 & 80.4 & 84.2 & 53.4 & 73.7 & 73.0 & 81.3 & 72.9 \\
\hline 
Faster R-CNN \cite{Faster} & 07+12 & 80.4 & 86.0 & 87.2 & 78.7 & 71.6 & 70.9 & 85.7 & 88.0 & 88.3 & 66.4 & 86.0 & 72.3 & 87.3 & 86.8 & 84.8 & 85.6 & 56.0 & 83.2 & 78.6 & 85.0 & 80.4 \\
Faster R-CNN+MPM & 07+12 & 81.6 & 87.4 & 87.4 & 85.1 & 71.7 & 72.1 & 86.9 & 88.6 & 89.1 & 67.6 & 87.5 & 75.1 & 88.5 & 87.8 & 85.4 & 86.3 & 58.8 & 85.7 & 78.7 & 86.0 & 78.4 \\
Faster R-CNN+MPM+CLM & 07+12 & 82.2 & 87.2 & 82.7 & 85.3 & 72.0 & 72.8 & 87.5 & 88.8 & 89.6 & 67.7 & 88.5 & 74.8 & 88.4 & 88.1 & 86.8 & 86.7 & 60.0 & 86.0 & 79.5 & 87.3 & 79.2 \\
\hline 
YoLo v3 \cite{YOLOv3} & 07 & 72.6 & 80.2 & 82.0 & 72.8 & 58.8 & 63.8 & 79.2 & 86.9 & 81.9 & 57.0 & 70.1 & 63.6 & 79.7 & 82.9 & 77.4 & 84.5 & 49.8 & 68.7 & 67.2 & 75.9 & 70.2 \\
YoLo v3+MPM+CLM & 07 & 74.5 & 76.9 & 84.7 & 75.2 & 57.4 & 60.7 & 81.8 & 82.8 & 83.7 & 58.1 & 78.8 & 67.1 & 82.5 & 85.4 & 80.7 & 85.0 & 48.8 & 72.7 & 71.2 & 81.4 & 75.6 \\
\hline 
YoLo v3 \cite{YOLOv3} & 07+12 & 78.5 & 83.6 & 89.2 & 78.6 & 60.9 & 64.4 & 85.8 & 89.2 & 87.0 & 60.7 & 79.5 & 74.7 & 85.8 & 88.1 & 83.7 & 88.8 & 55.5 & 75.4 & 76.6 & 85.3 & 77.7 \\
YoLo v3+MPM+CLM & 07+12 & 80.9 & 88.2 & 88.2 & 78.8 & 65.2 & 67.0 & 91.4 & 90.3 & 89.8 & 60.5 & 82.5 & 76.2 & 90.2 & 92.5 & 86.6 & 88.7 & 57.6 & 80.1 & 80.6 & 86.4 & 76.8 \\
\hline 
CenterNet \cite{CenterNet} & 07+12 & 75.3 & 82.8 & 82.6 & 73.2 & 58.2 & 61.6 & 81.6 & 84.3 & 86.4 & 60.7 & 76.3 & 69.5 &  83.3 & 86.5 & 81.6 & 83.0 & 51.8 & 73.8 & 73.3 & 79.7 & 75.2 \\
CenterNet+MPM+CLM & 07+12 & 76.9 & 81.1 & 88.3 & 78.1 & 60.3 & 65.7 & 82.3 & 89.2 & 84.7 & 62.0 & 79.9 & 69.9  & 81.8 & 85.7 & 82.2 & 87.8 & 54.5 & 74.6 & 73.1 & 80.3 & 75.5 \\
\end{tabular}}
\label{tab:detection}
\end{table*}

\begin{figure*}[t]
\centering
\includegraphics[width=\linewidth]{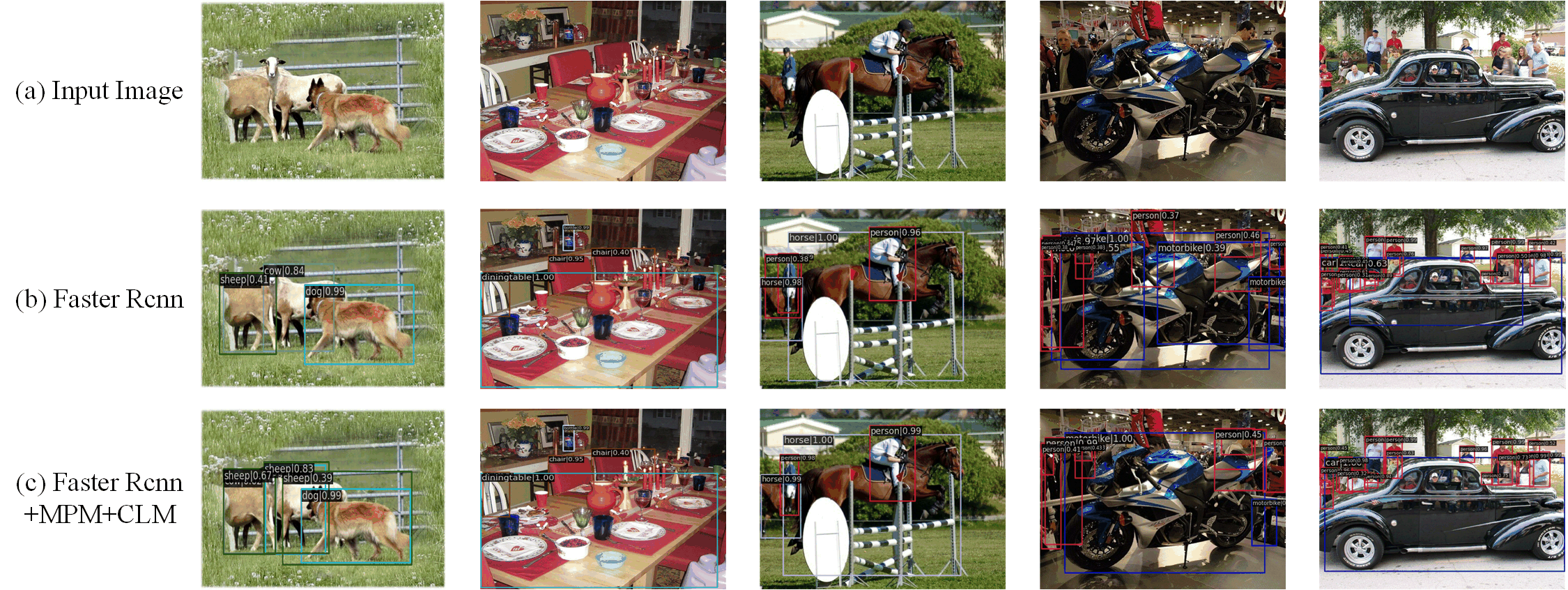}
\caption{Selected examples of object detection results on the PASCAL VOC 2007 test set using the Faster R-CNN and ``Faster R-CNN+MPM+CLM''.}
\label{fig:detection}
\end{figure*}

\subsection{\wqz{Flexibility of MPM and CLM}}

Smooth density map faces challenges in localizing crowded persons. Optimized maps is proposed to overcome the indistinguishable challenges in dense scenarios \cite{IIM, p2pnet, FIDT, autoscale, GMS}. AutoScale \cite{autoscale} proposes a simple and effective Learning to Scale (L2S) module, which automatically scales dense regions into reasonable closeness levels (reflecting image-plane distance between neighboring people). GMS \cite{GMS} proposes Gaussian Mixture Scope to regularize the chaotic scale distribution. \cite{IIM} proposed an effective end-to-end Independent Instance Maps segmentation (IIM) to locate head positions in crowd scenes. P2PNet \cite{p2pnet} directly predicts a set of point proposals to represent heads in an image, being consistent with the human annotation results. \cite{FIDT} proposes a novel Focal Inverse Distance Transform (FIDT) map for the crowd localization task. The proposed LDFNet including MPM and CLM is flexible, incorporating them into existing models can potentially boost their localization performances in dense scenarios. We incorporate the proposed LDFNet into these methods and conduct experiments on UCF-QNRF and NWPU dataset. For IIM, the proposed LDFNet is applied in the ``predictor'' network. For P2PNet, we plug MPM into the regression head which is used for localization and plug CLM into the classification head which is used for classification. For FIDT, we incorporate LDFNet into the ``regressor'' network. The localization performances are presented in Table \ref{localization_Q} and \ref{localization_N1}. 

Following the previous crowd localization challenge, NWPU-Crowd \cite{NWPU}, instance-level Precision, Recall and F1-measure are utilized to evaluate models. When the distance between the given predicted point and ground truth point is less than a distance threshold $\sigma$, it means the predicted and ground truth point are successfully matched. The localization metrics of different methods are different. FIDT and AutoScale choose a fix thresholds ($\sigma=8$) for evaluation in Shanghai Part A and B. IIM chooses ($\sigma=\sqrt{w^2+h^2}/2$) for evaluation in Shanghai Part A and B, where $w$ and $h$ are the width and height of the instance, respectively. Specifically, IIM exploits the box labels provided by NWPU-Crowd to train a head scale prediction network, which automatically generates box-level annotation ($w$ and $h$) for the datasets only with point annotation. The remaining experimental settings are consistent with those reported in the original paper.

The results presented in Table \ref{localization_Q} demonstrate that our proposed "FIDT+LDFNet" model surpasses the performance of other methods in the UCF-QNRF dataset. Similarly, for the NWPU dataset, as shown in Table \ref{localization_N1}, the "IIM+LDFNet" model achieves unparalleled performance in Precision and F1-measure categories. We have conducted localization experiments on Shanghai Part datasets. The related results have been presented in TABLE \ref{localization_SH}. We observe that the integration of LDFNet significantly improves the performance of the original methods across all localization metrics. It is noteworthy that the integration of LDFNet enhances the performance of the original methods across all localization metrics. This improvement can be attributed to the proposed MPM module, which, through its masking-reconstructing mechanism, enables the model to comprehend the contents of the masked regions. Additionally, our proposed CLM module differentiates the foreground and background in the feature space, thereby reducing confusion in high-density regions. These two modules collectively enable the model to learn distinguishable features, which ultimately aids in better localization. These findings underscore the flexibility of the proposed LDFNet and suggest that its integration into optimized maps (such as IIM, P2PNet, and FIDTM) can lead to improved performances.

% For UCF-QNRF, the results presented in Table \ref{localization_Q} show that our proposed ``FIDT+LDFNet'' outperform other methods. For localization results of NWPU in Table \ref{localization_N1}, ``IIM+LDFNet'' achieves the best performance in Precision and F1-measure. One could be observed that incorporating LDFNet performs better than the original methods on each localization metric. The reason is that the proposed MPM with masking-reconstructing mechanism allows the model to learn about what is present in the masked regions. And the proposed CLM separates foreground and background in feature space, thus mitigating the confusion in high-density regions. These two modules allow the model to learn distinguishable features, benefiting localization. These results indicate that the proposed LDFNet is flexible and employing LDFNet into optimized maps (IIM, P2PNet, and FIDTM) would achieve better performances. 

    \begin{table*}[htbp]
\caption{Localization performance on UCF-QNRF dataset.}
\begin{center}
\begin{tabular}{llllll}
\hline & Precision & Recall & F1-measure & MAE & RMSE\\
\hline 
MCNN \cite{MCNN} & $0.599$ & $0.635$ & $0.617$ & 277.0 & 426.0\\
CL \cite{CL} & $0.758$ & $0.598$ & $0.668$ & 132.0 & 191.0\\
LCFCN \cite{LCFCN} & $0.779$ & $0.524$ & $0.627$ & - & -\\
GL \cite{GL} & ${0.782}$ & ${0.748}$ & ${0.763}$ & 84.3 & 147.5\\
TopoCount \cite{topocount} & $0.818$ & $0.790$ & $0.822$ & 89.0 & 159.0\\
LSC-CNN \cite{LSC-CNN} & $0.746$ & $0.735$ & $0.741$ & 225.6 & 302.7\\
\hline
IIM \cite{IIM} & $0.793$ & $0.659$ & $0.720$ & - & -\\ 
IIM+LDFNet (ours) & $0.798$ \textcolor{red}{(+0.63\%)} & $0.676$ \textcolor{red}{(+2.57\%)} & $0.732$ \textcolor{red}{(+1.66\%)} & - & -\\ 
P2PNet \cite{p2pnet} & $0.712$ & $0.758$ & $0.721$ & 85.3 & 154.5 \\
P2PNet+LDFNet (ours) & $0.738$ \textcolor{red}{(+3.65\%)} & $0.779$ \textcolor{red}{(+2.77\%)} & $0.758$ \textcolor{red}{(+5.13\%)} & 80.2 \textcolor{red}{(+5.98\%)} & 141.8 \textcolor{red}{(+8.22\%)}\\ 
FIDT \cite{FIDT} & $0.845$ & $0.801$ & $0.822$ & 89.0 & 153.5\\
FIDT+LDFNet (ours) & $\textbf{0.851}$ \textcolor{red}{(+0.71\%)}& $\textbf{0.820}$ \textcolor{red}{(+2.37\%})& $\textbf{0.829}$ \textcolor{red}{(+0.85\%)} & 83.5 \textcolor{red}{(+6.18\%)} & 143.7 \textcolor{red}{(+6.38\%)}\\ 
\hline
\end{tabular}
\label{localization_Q}
\end{center}
\end{table*}

\begin{table*}[htbp]
\caption{Localization performance on NWPU-Crowd dataset.}
\begin{center}
\begin{tabular}{llllll}
\hline Method & Precision & Recall & F1-measure & MAE & RMSE\\
\hline 
% Faster RCNN \cite{Faster} & $\textbf{0.958}$ & $0.035$ & $0.068$ \\
TinyFace \cite{tiny} & $0.529$ & ${0.611}$ & $0.567$ & - & -\\
VGG+GPR & $0.558$ & $0.496$ & $0.525$ & - & -\\
RAZNet \cite{RAZNet} & $0.666$ & $0.543$ & ${0.599}$ & - & -\\
D2CNet \cite{D2CNet} & $0.729$ & $0.662$ & ${0.700}$ & 85.5 & 361.5\\
TopoCount \cite{topocount} & $0.695$ & ${0.687}$ & $0.691$ & 107.8 & 438.5\\
CrossNet-HR \cite{CrossNet} & $0.748$ & ${0.757}$ & ${0.739}$ & 70.4 & 325.1\\
GL \cite{GL} & ${0.800}$ & ${0.562}$ & ${0.660}$ & 76.8 & 343.0\\
\hline
IIM \cite{IIM} & $0.813$ & $0.717$ & $0.762$ & 87.1 & 406.2\\ 
IIM+LDFNet (ours) & $\textbf{0.841}$ \textcolor{red}{(+3.44\%)}& $0.723$ \textcolor{red}{(+0.83\%)}& $\textbf{0.778}$ \textcolor{red}{(+2.10\%)} & 81.2 \textcolor{red}{(+6.77\%)}& 376.6 \textcolor{red}{(+7.29\%)}\\ 
P2PNet \cite{p2pnet} & $0.729$ & $0.695$ & $0.712$ & 77.4 & 362.0\\
P2PNet+LDFNet (ours) & $0.758$ \textcolor{red}{(+3.98\%)}& $\textbf{0.781}$ \textcolor{red}{(+12.37\%)}& $0.769$ \textcolor{red}{(+8.00\%)} & 72.4 \textcolor{red}{(+6.46\%)}& 341.8 \textcolor{red}{(+5.58\%)}\\ 
FIDT \cite{FIDT} & $0.797$ & $0.717$ & $0.755$ & 86.0 & 312.5\\
FIDT+LDFNet (ours) & $0.814$ \textcolor{red}{(+2.13\%)}& $0.724$ \textcolor{red}{(+0.98\%)}& $0.767$ \textcolor{red}{(+1.59\%)} & 78.4 \textcolor{red}{(+8.84\%)} & 301.4 \textcolor{red}{(+3.55\%)}\\ 
\hline
\end{tabular}
\label{localization_N1}
\end{center}
\end{table*}

    \begin{table*}[htbp]
\caption{Localization performance on Shanghai Part dataset.}
\begin{center}
\begin{tabular}{lllllll}
\hline \multirow{2}{*} {Method} & \multicolumn{3}{c}{Shanghai Part A} & \multicolumn{3}{c}{Shanghai Part B} \\
& Precision & Recall & F1-measure & Precision & Recall & F1-measure\\
\hline 
% Faster RCNN \cite{Faster} & $\textbf{0.958}$ & $0.035$ & $0.068$ \\
TinyFace \cite{tiny} & $0.431$ & $0.855$ & $0.573$ & $0.647$ & $0.790$ & $0.711$\\
RAZNet \cite{RAZNet} & $0.613$ & $0.795$ & ${0.692}$ & $0.600$ & $0.783$ & $0.680$\\
LSC-CNN \cite{LSC-CNN} & $0.696$ & $0.665$ & $0.680$ & $0.717$ & $0.706$ & $0.712$\\
AutoScale \cite{autoscale} & $0.744$ & $0.717$ & $0.730$ & $-$ & $-$ & $-$\\
GMS \cite{GMS} & $0.817$ & $0.749$ & $0.781$ & $0.919$ & $0.812$ & $0.863$\\
\hline
IIM \cite{IIM} & $0.763$ & $0.705$ & $0.733$ & $0.898$ & $0.786$ & $0.838$\\ 
IIM+LDFNet (ours) & $0.775$ \textcolor{red}{(+1.57\%)}& $0.724$ \textcolor{red}{(+2.70\%)}& $0.749$ \textcolor{red}{(+2.18\%)}& $0.908$ \textcolor{red}{(+1.11\%)}& $0.794$ \textcolor{red}{(+1.04\%)}& $0.847$ \textcolor{red}{(+1.07\%)}\\  
FIDT \cite{FIDT} & $0.782$ & $0.770$ & $0.776$ & $0.839$ & $0.832$ & $0.835$\\
FIDT+LDFNet (ours) & $0.789$ \textcolor{red}{(+0.89\%)}& $0.780$ \textcolor{red}{(+1.30\%)}& $0.785$ \textcolor{red}{(+1.16\%)}& $0.848$ \textcolor{red}{(+1.07\%)}& $0.840$ \textcolor{red}{(+0.96\%)}& $0.844$ \textcolor{red}{(+1.08\%)}\\ 
\hline
\end{tabular}
\label{localization_SH}
\end{center}
\end{table*}

% \begin{table}[htbp]
%     \centering
%     \begin{tabular}{c|c|c|c}
%     \hline
%         \multirow{2}{*} {Method} & \multirow{2}{*} {Label} & {UCF-QNRF} & {NWPU} \\
%         & & {F1-m/Pre/Rec (\%)} & {F1-m/Pre/Rec (\%)} \\
%     \hline
        
%     \end{tabular}
%     \caption{Caption.}
%     \label{tab:localization}
% \end{table}

The proposed MPM and CLM can also benefit object detection, where cluttered environments pose challenges to accurate localization and recognition. We adopt MMDetection \cite{MMDetection}, an object detection toolbox that contains a rich set of object detection methods, to conduct related experiments on PASCAL-VOC 07 \cite{voc07}. Specifically, we integrate the proposed MPM and CLM into three classical object detection methods: the two-stage detector --- Faster R-CNN \cite{Faster}, one-stage detector --- YoLo v3 \cite{YOLOv3}, and anchor-free detector --- CenterNet \cite{CenterNet}. Following MMDetection, we divide the object detection framework into three parts (backbone, neck, and head). The backbones of Faster R-CNN \cite{Faster}, YoLo v3 \cite{YOLOv3}, and anchor-free detector are ResNet-50, DarkNet-53, and ResNet-18. The backbone in the object detection framework is equivalent to the backbone of our counting model, and the head is equivalent to our regression decoder. As a result, we put MPM on the latest layer of the backbone and put CLM between the neck and head. The detailed results are presented in Table~\ref{tab:detection}. 

We find that using MPM or CLM can significantly improve the performance of object detection methods, e.g., plugging MPM/CLM into Faster R-CNN improves mAP from 72.5 to 74.2/73.5. To demonstrate the effectiveness of our proposed mask-prediction mechanism, we add Transformer to Faster R-CNN. Comparing the experimental results of ``Faster R-CNN+Transformer''(72.8) and ``Faster R-CNN+MPM''(74.2), we can find that the accuracy improvement brought by MPM is mainly due to the mask-prediction mechanism. Such mask-prediction mechanism can help model know the content of the masked regions and contribute to improved localization capabilities, which is effective for object detection. Combining MPM and CLM could further boost the detection accuracy of different detection methods, e.g. plugging MPM and CLM into Faster R-CNN improves mAP from 72.5/80.4 to 75.1/82.2. Therefore, we can conclude that the proposed MPM and CLM are flexible and can be applied to existing detection models, significantly outperforming the baselines.

To further show the effectiveness of the proposed MPM and CLM, we visualize the object detection results on PASCAL-VOC 07 \cite{voc07} test set using the Faster R-CNN and ``Faster R-CNN+MPM+CLM'' in Figure \ref{fig:detection}. From Figure \ref{fig:detection}, we can see more accurate localization boxes are obtained by the proposed ``Faster R-CNN+MPM+CLM''. For example, in the 4-th column, Faster Rcnn produces redundant boxes when detecting the motorbike. In contrast, our ``Faster R-CNN+MPM+CLM'' contributes to improve localization quality of detected boxes, among which redundant ones are easy to be suppressed.

% \begin{table*}[htbp]
%     \centering
%     \caption{Flexibility of MPM and CLM. We add MPM and CLM into different detection frameworks on pascal-voc 07 \cite{voc07} dataset.}
%     \setlength{\tabcolsep}{10mm}{\begin{tabular}{cccccc}
%     \hline Model & Data & Resolution & Backbone & $\textbf{A} \textbf{P}^{\text {50}}$ \\
%     \hline 
%     Faster R-CNN \cite{Faster} & 07 & 600$\times$1000 & ResNet-50 & 72.5\\
%     Faster R-CNN+Transformer & 07 & 600$\times$1000 & ResNet-50 & 72.8\\
%     Faster R-CNN+CLM & 07 & 600$\times$1000 & ResNet-50 & 73.5\\
%     Faster R-CNN+MPM & 07 & 600$\times$1000 & ResNet-50 & 74.2\\
%     Faster R-CNN+MPM+CLM & 07 & 600$\times$1000 & ResNet-50 & 75.1\\
%     \hline
%     Faster R-CNN & 07+12 & 600$\times$1000 & ResNet-50 & 80.4\\
%     Faster R-CNN+MPM & 07+12 & 600$\times$1000 & ResNet-50 & 81.6\\
%     Faster R-CNN+MPM+CLM & 07+12 & 600$\times$1000 & ResNet-50 & 82.2\\
%     \hline
%     YoLo v3 \cite{YOLOv3} & 07 & 320$\times$320 & DarkNet-53 & 72.7\\
%     YoLo v3+MPM+CLM & 07 & 320$\times$320 & DarkNet-53 & 74.6\\
%     \hline
%     YoLo v3 & 07+12 & 320$\times$320 & DarkNet-53 & 78.5\\
%     YoLo v3+MPM+CLM & 07+12 & 320$\times$320 & DarkNet-53 & 80.8\\
%     \hline
%     CenterNet \cite{CenterNet} & 07+12 & 512$\times$512 & ResNet-18 & 75.4\\
%     CenterNet+MPM+CLM & 07+12 & 512$\times$512 & ResNet-18 & 76.8\\ 
%     \hline
%     \end{tabular}}
    
%     \label{tab:detection}
% \end{table*}

\subsection{Complexity analysis.}

\begin{table}[htbp]
\centering
\caption{Comparison of the Parameters (M), FLOPs (G), and Inference speed (s / 100 images).}
\setlength{\tabcolsep}{4mm}{\begin{tabular}{cccc}
\hline
Model & Parameters & FLOPs & Inference speed \\
\hline
DM-Count \cite{DM-Count} & $21.5$ & $56.9$ & $3.1$\\
+CLM & $22.2$ & $59.8$ & $3.1$\\
+Trans & $29.9$ & $58.0$ & $3.8$\\
+MPM & $30.9$ & $58.5$ & $3.9$\\
+CLM+MPM & $31.6$ & $61.4$ & $3.9$\\
ViT-B \cite{ViT} & $86.0$ & $55.4$ & $8.8$\\
\hline
\end{tabular}}
\label{tab:complexity}
\end{table}

Table \ref{tab:complexity} reports a comparison of model size and floating point operations (FLOPs) computed on one 384 × 384 input image with a single NVIDIA 3090 GPU. And the inference time is the average time of 100 runs on testing 1024×1024 sample. It can be easily observed that the proposed CLM hardly increases the computational overhead. And the model size and inference speed of our LDFNet (+CLM+MPM) are much smaller than those of ViT-B \cite{ViT}. It shows the proposed components are lightweight compared with vanilla transformers. Though our full model has more learnable parameters and FLOPS, it is a real-time model, processing 25.6 images per second.

\section{Conclusion}
In this paper, we propose LDFNet, a framework designed to learn discriminative features for crowd counting, while addressing the problems of localization and confusion on foreground and background that plague existing regression-based models. To address the problems of localization, we propose a masked feature prediction module that randomly masks feature vectors in the feature map and then reconstructs them. This module allows the model to learn about what is present in the masked regions, thereby improving the model’s ability to localize objects in high-density regions. To mitigate the confusion on foreground and background, we introduce a supervised pixel-level contrastive learning module that pulls targets close to each other and pushes them far away from background in the feature space. This module could separate foreground and background in feature space, thus mitigating the confusion in high-density regions. The proposed MPM and CLM are flexible and can be applied to existing counting and detection models, significantly outperforming the baselines.

\ifCLASSOPTIONcompsoc
  % The Computer Society usually uses the plural form
  \section*{Acknowledgments}
\else
  % regular IEEE prefers the singular form
  \section*{Acknowledgment}
\fi
This work was supported by the National Key Research and Development Program of China under Grant No. 2020AAA0108100,  the National Natural Science Foundation of China under Grant No. 62073257, 62141223 and the Key Research and Development Program of Shaanxi Province of China under Grant No. 2022GY-076.

% Can use something like this to put references on a page
% by themselves when using endfloat and the captionsoff option.
\ifCLASSOPTIONcaptionsoff
  \newpage
\fi

% trigger a \newpage just before the given reference
% number - used to balance the columns on the last page
% adjust value as needed - may need to be readjusted if
% the document is modified later
%\IEEEtriggeratref{8}
% The "triggered" command can be changed if desired:
%\IEEEtriggercmd{\enlargethispage{-5in}}

% references section

% can use a bibliography generated by BibTeX as a .bbl file
% BibTeX documentation can be easily obtained at:
% http://mirror.ctan.org/biblio/bibtex/contrib/doc/
% The IEEEtran BibTeX style support page is at:
% http://www.michaelshell.org/tex/ieeetran/bibtex/
\bibliographystyle{IEEEtran}
% argument is your BibTeX string definitions and bibliography database(s)
% \bibliography{IEEEabrv,../bib/paper}
\bibliography{egbib}
\end{document}